\documentclass[journal,twoside]{IEEEtran}
\IEEEoverridecommandlockouts
\usepackage{graphicx} \graphicspath{ {figures/} }
\usepackage{amsmath,amssymb,mathabx,amsbsy,mathtools,etoolbox}
\usepackage[utf8]{inputenc} % allow utf-8 input
\usepackage[T1]{fontenc}    % use 8-bit T1 fonts

\usepackage{algorithmic}
\usepackage[ruled]{algorithm2e}
\usepackage{acronym}
\usepackage{enumitem}
\usepackage[breaklinks,colorlinks,linkcolor=black]{hyperref}
\usepackage{balance}
\usepackage{xspace}
\usepackage{setspace}
\usepackage[skip=3pt,font=small]{subcaption}
\usepackage[skip=3pt,font=small]{caption}
\usepackage[dvipsnames,svgnames,x11names]{xcolor}
\usepackage[capitalise,nameinlink]{cleveref}
\usepackage{booktabs,tabularx,colortbl,multirow,multicol,array,makecell}
\usepackage{pifont}
\usepackage{cite}
\usepackage{tikz}
\usepackage{nicematrix}
\makeatletter
\DeclareRobustCommand\onedot{\futurelet\@let@token\@onedot}
\def\@onedot{\ifx\@let@token.\else.\null\fi\xspace}

\def\etal{\emph{et al}\onedot}
\makeatother

\makeatletter
\def\BState{\State\hskip-\ALG@thistlm}
\makeatother

\makeatletter
\renewcommand{\paragraph}{%
  \@startsection{paragraph}{4}%
  {\z@}{0ex \@plus 0ex \@minus 0ex}{-1em}%
  {\hskip\parindent\normalfont\normalsize\bfseries}%
}
\makeatother

\crefname{algorithm}{Alg.}{Algs.}
\Crefname{algocf}{Algorithm}{Algorithms}
\crefname{section}{Sec.}{Secs.}
\Crefname{section}{Section}{Sections}
\crefname{table}{Tab.}{Tabs.}
\Crefname{table}{Table}{Tables}
\crefname{figure}{Fig.}{Fig.}
\Crefname{figure}{Figure}{Figure}
\definecolor{gblue}{HTML}{4285F4}
\definecolor{gred}{HTML}{DB4437}
\definecolor{ggreen}{HTML}{0F9D58}

\definecolor{mygray}{gray}{.92}

\acrodef{qp}[QP]{Quadratic Programming}
\acrodef{fd}[FD]{Force Decomposition}
\acrodef{ros}[ROS]{Robot Operating System}
\acrodef{uav}[UAV]{Unmanned Aerial Vehicle}
\acrodef{dof}[DoF]{Degree-of-freedom}
\acrodef{com}[CoM]{Center-of-Mass}
\acrodef{ftc}[FTC]{fault-tolerant control}
\begin{document}

\title{Fault-tolerant Control of an Over-actuated UAV Platform Built on Quadcopters and Passive Hinges} 
\author{Yao Su,~\IEEEmembership{Member,~IEEE}, Pengkang Yu, Matthew J. Gerber, Lecheng Ruan \\and Tsu-Chin Tsao,~\IEEEmembership{Senior Member,~IEEE}

\thanks{Manuscript received: 2 August 2022; revised 20 February 2023 and 24 April 2023; accepted: 12 May, 2023. Date of publication xxxx, 2023; date of current version xxxx, 2023.  Recommended by Technical Editor xxx and Senior Editor Kostas Kyriakopoulos. (\textit{Yao Su and Pengkang Yu contributed equally to this work.}) \textit{(Corresponding authors: Yao Su and Lecheng Ruan.)}}

\thanks{Yao Su, Pengkang Yu, Matthew J. Gerber, Lecheng Ruan, and  Tsu-Chin Tsao are with Mechanical and Aerospace Engineering Department, University of California, Los Angeles (UCLA), Los Angeles, CA
90095 USA (e-mail: yaosu@ucla.edu; paulyu1994@ucla.edu; gerber211@ucla.edu; ruanlecheng@ucla.edu; ttsao@ucla.edu).}
%\thanks{The video of experiments is available at \url{https://www.youtube.com/watch?v=Y-8l1EBH5gU}.}
\thanks{This article has supplementary material provided by the authors and color versions of one or more figures available at https://doi.org/10.1109/TMECH.2023.}
\thanks{Digital Object Identifier }}

\markboth{IEEE/ASME TRANSACTIONS ON MECHATRONICS}
{Su \MakeLowercase{\textit{et al.}}: Fault-tolerant Control of an Over-actuated UAV Platform Built on Quadcopters and Passive Hinges} 

\maketitle

\begin{abstract}
Propeller failure is a major cause of multirotor \acp{uav} crashes. While conventional multirotor systems struggle to address this issue due to underactuation, over-actuated platforms can continue flying with appropriate \ac{ftc}. This paper presents a robust \ac{ftc} controller for an over-actuated \ac{uav} platform composed of quadcopters mounted on passive joints, offering input redundancy at both the high-level vehicle control and the low-level quadcopter control of vectored thrusts. To maximize the benefits of input redundancy during propeller failure, the proposed \ac{ftc} controller features a hierarchical control architecture with three key components: (i) a low-level adjustment strategy to prevent propeller-level thrust saturation; (ii) a compensation loop for mitigating introduced disturbances; (iii) a nullspace-based control allocation framework to avoid quadcopter-level thrust saturation. Through reallocating actuator inputs in both the low-level and high-level control loops, the low-level quadcopter control can be maintained with up to two failed propellers, ensuring that the whole platform remains stable and avoids crashing. The proposed controller's superior performance is thoroughly examined through simulations and real-world experiments.
\end{abstract}
\begin{IEEEkeywords}
   Over-actuated UAV, propeller failure, fault-tolerant control (FTC), nullspace allocation, input redundancy, optimization
\end{IEEEkeywords}

\section{Introduction}
\IEEEPARstart{O}{ver-actuated} multirotor \ac{uav}s have been proposed in the last decade to overcome the underactuation issue of traditional co-linear multirotor UAVs~\cite{qian2020guidance}, leveraging vectored thrusts to improve dynamic properties. There are mainly two categories of realizations in this field. The first group of works~\cite{kamel2018voliro,rajappa2015modeling,saied2015fault} employs multiple propeller-motor pairs in various or varying directions to achieve full or over-actuation. The second group of works~\cite{nguyen2018novel,yu2021over, ruan2020independent, pi2021simple, ruan2023control} utilizes standard quadcopters mounted on passive joints as actuation modules, simplifying the design and prototyping process while reducing internal disturbance levels~\cite{ruan2023control}. With additional actuators onboard, these platforms should demonstrate increased robustness against \textbf{propeller failure} compared to conventional quadcopters: without sufficient \ac{ftc} algorithms, over-actuated multirotor systems are more likely to suffer from propeller failure than quadcopters, resulting in crashes. Therefore, developing \ac{ftc} algorithms to reduce crash likelihood in the event of propeller failure is of great interest.

\begin{figure}[t!] 
    \centering
    \includegraphics[width=\linewidth,trim=3.5cm 5cm 11.5cm 2cm,clip]{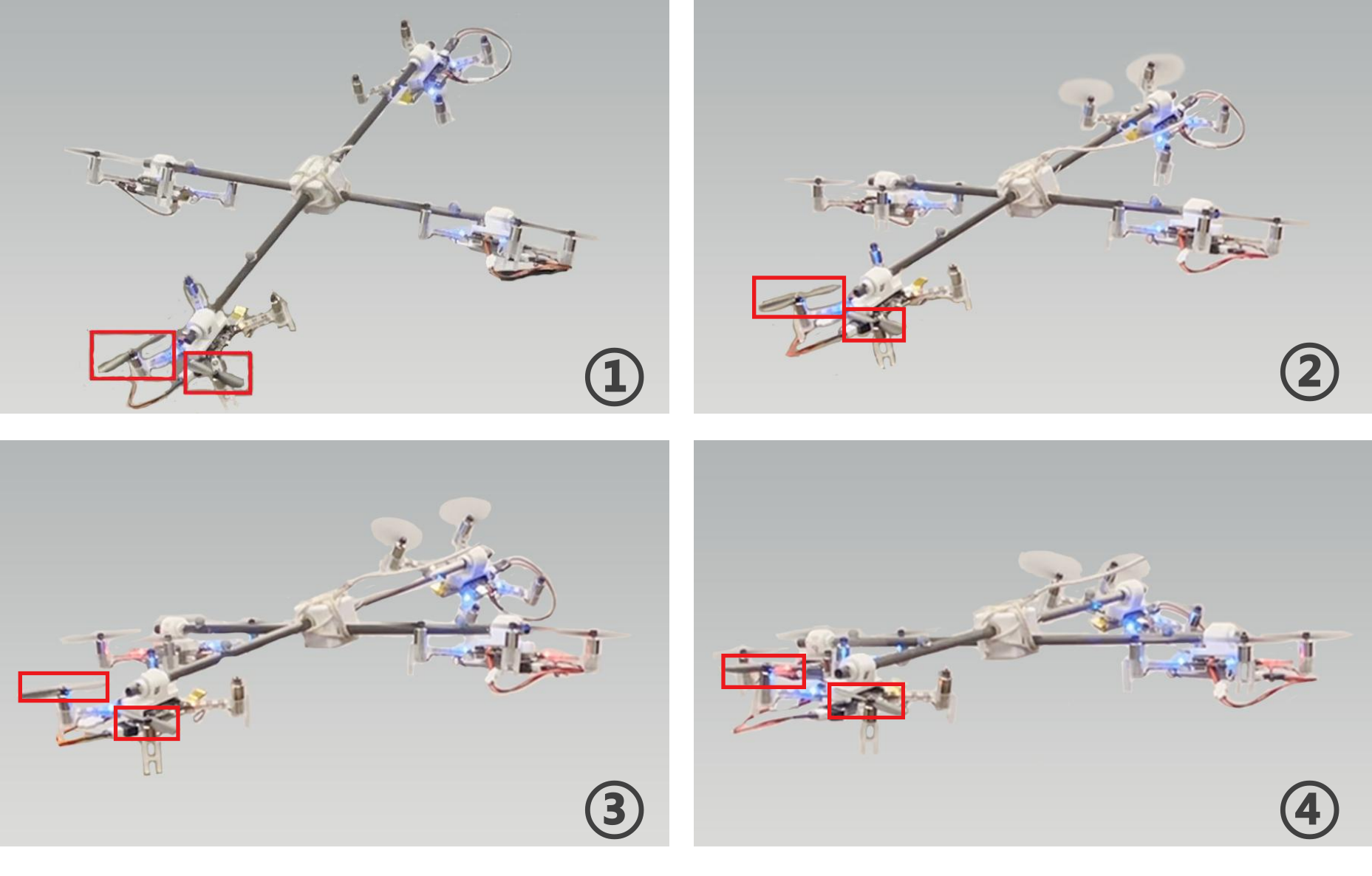}
    \caption{\textbf{With two failed propellers on one quadcopter module, our proposed \ac{ftc} algorithm can prevent the platform from crashing and maintain the capability of trajectory tracking.} (Failed propellers are labeled by red boxes).}
    \label{fig:clips}
\end{figure}

In this paper, we implement the previously proposed nullspace-based control allocation framework~\cite{su2021nullspace} to address the \textbf{thrust force saturation} issue in over-actuated \ac{uav} platforms. We validate this approach on our customized over-actuated platform~\cite{pi2021simple}, where four mini-quadcopters are connected to the mainframe via 1 \ac{dof} passive hinges, serving as tiltable thrust generators. The platform features input redundancy in both high-level wrench control and low-level tiltable thrusts control~\cite{su2021fast}. Following this, we propose a \ac{ftc} algorithm specifically for scenarios in which one or more propellers on a single quadcopter (denoted as \textbf{Bad QC}) fail while the other quadcopters (denoted as \textbf{Good QCs}) remain functional (see \cref{fig:clips}). 
This \ac{ftc} controller employs a hierarchical structure with three main components: (i) a low-level controller to adjust the propeller-level thrust force distribution on the Bad QC, (ii) a high-level controller to reallocate quadcopter-level thrust force distribution among the Good and Bad QCs, and (iii) a compensation loop for disturbance attenuation.

In the low-level controller, due to the reduced thrust and torque capacity, the thrust distribution among the propellers of the Bad QC is adjusted to maintain control of the tilting angle and the thrust. However, this low-level adjustment introduces an interaction torque between the central frame and the Bad QC, acting as a disturbance. 
To attenuate this undesirable torque, the redundant inputs of the three Good QCs are utilized to formulate a compensation loop~\cite{su2021fast}. 
Finally, in the high-level controller, the nullspace-based allocation framework is implemented to optimize the thrust distribution of all four quadcopters. This control framework is validated through both dynamic simulations and real-world experiments. 

Our contributions are highlighted as follows:
\begin{itemize}
\item[(1)] We analyze the thrust force saturation issue in over-actuated \ac{uav} platforms, and implement the nullspace-based control allocation framework to address it. 
\item[(2)] We develop a \ac{ftc} algorithm to fully utilize the redundancy of over-actuated \ac{uav} when some of the propellers on a single quadcopter module fail. We analyze and compare two different low-level control methods, design a compensation loop for disturbance attenuation, and incorporate a nullspace-based control allocation to find the optimal allocation solution under this scenario.  
\item[(3)] We present simulation and experimental validations to demonstrate the effectiveness of our proposed control algorithm in handling the thrust force saturation and propeller failure in over-actuated \ac{uav} platforms.
\end{itemize}

The remainder of the paper is organized as follows. 
Related work is summarized in \cref{sec:related}. 
The dynamics models of the over-actuated UAV platform are reviewed in \cref{sec:dynamics and review}. 
\cref{sec:control} presents the control architecture and the nullspace-based framework for control allocation. 
\cref{sec:Robust control} describes the \ac{ftc} framework to handle propeller failure. 
\cref{sec:simulation} and \cref{sec:experiment} present the simulation and experiment results. 
Finally, we conclude the paper with a discussion in \cref{sec:discussion} and \cref{sec:conclusion}.

\section{Related Work} \label{sec:related}
\subsection{Control Allocation}
\textbf{Control allocation} in \textbf{over-actuated \ac{uav}} platforms, which computes individual actuator commands from the desired total wrench, is a constrained nonlinear optimization problem that is generally difficult to solve with high efficiency. Ryll \etal first utilized dynamic output linearization for control allocation at a higher differential level,  requiring accurate acceleration measurements or estimation~\cite{ryll2014novel}. Kamel \etal introduce a \ac{fd}-based method, transforming the nonlinear allocation problem into a linear one by defining intermediate variables~\cite{kamel2018voliro}. 
This method improved computational speed by directly choosing the least-square solution but sacrificing input redundancies. 
Furthermore, iterative approach~\cite{zhao2020enhanced} 
and separation method~\cite{santos2021fast} were proposed for improved efficiency.
However, none of these methods \cite{ryll2014novel,kamel2018voliro,zhao2020enhanced,santos2021fast} considered input constraints, leading to instability when the input constraints were triggered~\cite{johansen2013control}. 

The \ac{qp}-based framework~\cite{johansen2004constrained} relied on discretization and linearization to incorporate both inequality and equality constraints. Nonetheless, it only generated approximate solutions, introducing additional disturbance to the control system. In our previous work~\cite{su2021nullspace}, we developed a \textbf{nullspace-based} allocation framework to combine the benefits of \ac{fd}-based and \ac{qp}-based frameworks and provide exact allocation solutions that satisfied the defined input constraints in real-time. Specifically, we demonstrated its capability by addressing the kinematic-singularity problem of a twist-and-tilt rotor platform~\cite{yu2021over}. In this paper, we further implement this framework to address the issue of thrust force saturation and propeller failure on a different over-actuated UAV platform. 

\subsection{Fault-tolerant Control}
\ac{uav} \textbf{\acf{ftc}} strategies can be categorized based on their configurations.
For quadcopters, due to underactuation, propeller failure typically requires sacrificing yaw motion control to maintain full translational control~\cite{nguyen2020design,lee2020fail,chung2020fault,mueller2016relaxed,shao2021adaptive}. Multirotor \acp{uav} with more than six controllable inputs~\cite{michieletto2017control,du2015controllability,marks2012control,saied2015fault}, or tilt-rotor quadcopters~\cite{gerber2018twisting,li2021flexibly,ding2020tilting,park2018odar} exhibit greater robustness against propeller loss due to \textbf{input redundancy}. The Y-shaped hexarotor platform with tilted rotors exhibited enhanced rotor-failure robustness compared to the standard star-shaped hexarotor platform~\cite{michieletto2018fundamental}. Related \ac{ftc} controllers were presented in~\cite{nguyen2018fault} based on modifications of control allocation and in~\cite{pose2021multirotor} based on Center-of-Mass shifting.  A new type of over-actuated \ac{uav} platforms integrates quadcopters and passive joints to achieve full actuation~\cite{nguyen2018novel,pi2021simple,yu2021over}.
Inherently, these platforms possess more propellers than standard tilt-rotor platforms, increasing the likelihood of propeller failure. In this paper, we present an \ac{ftc} for this type of \ac{uav} platform that sufficiently utilizes the redundancy of the entire platform at both high-level and low-level control, enhancing platform robustness against propeller failure. As a result, various \ac{uav} platforms~\cite{gerber2018twisting,yu2021over,zhao2020enhanced,li2021flexibly,da2020drone,csenkul2016system}, which may have different thrust generation capabilities among propellers or thrust-generation modules under propeller failure, can achieve improved rotor-failure robustness.

\section{Platform Dynamics}
\label{sec:dynamics and review}
\begin{figure}[t!]
    \centering
    \includegraphics[width=0.9\linewidth,trim=6cm 1cm 5.8cm 1.5cm,clip]{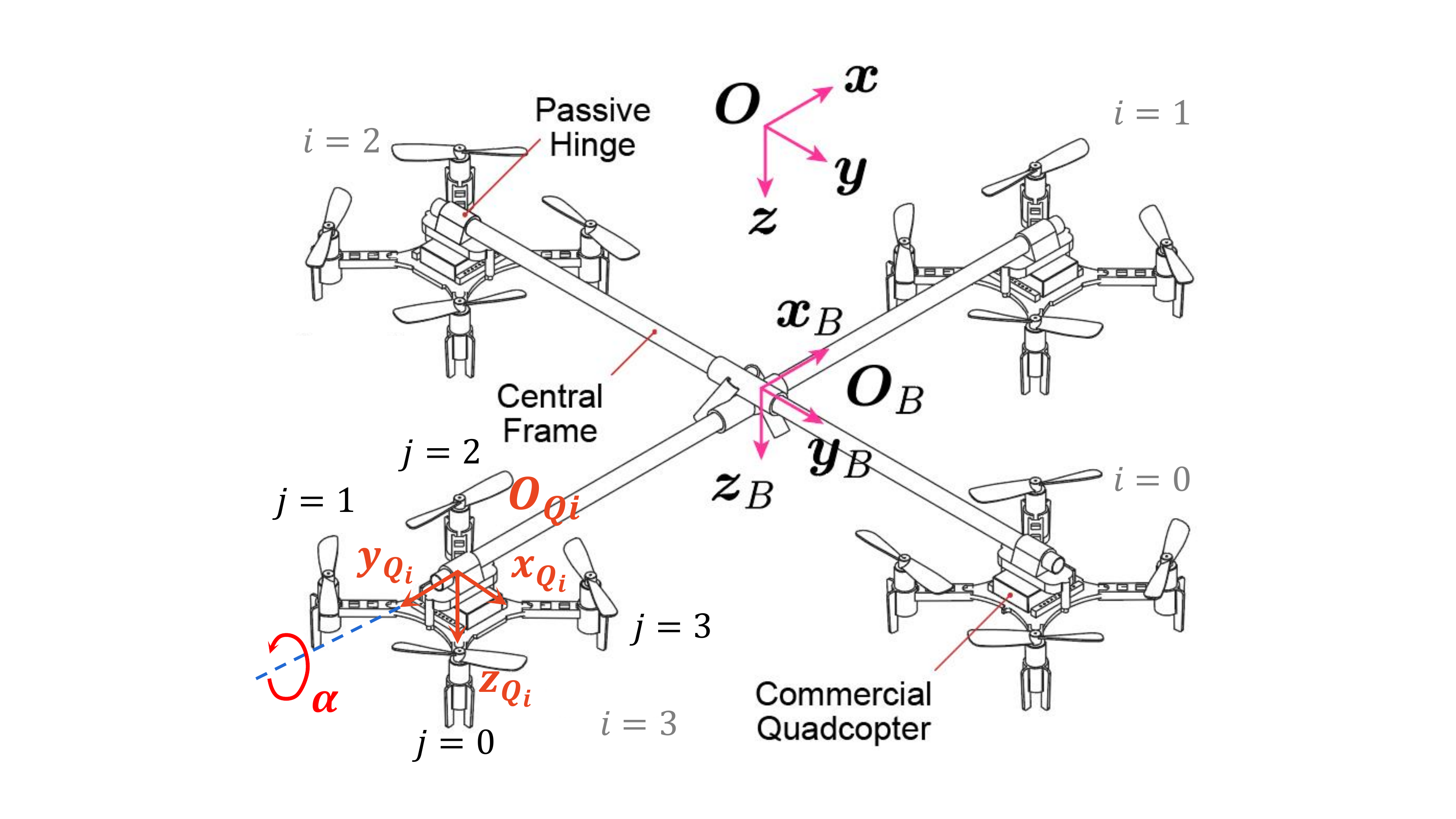}
    \caption{The platform presented in this paper: four commercial quadcopters are passively hinged to the central frame with 1 \ac{dof} connection.} 
    \label{fig:hinge}
\end{figure}

The over-actuated UAV discussed in this paper is built upon regular quadcopters mounted on passive hinges and is a representative configuration for over-actuated UAVs with auxiliary inputs~\cite{pi2021simple,ruan2023control}. As shown in \cref{fig:hinge}, we define the world frame, body frame, and quadcopter frames as $\mathcal{F}_W$, $\mathcal{F}_B$, 
and $\mathcal{F}_{Q_i}$, respectively. The position of the central frame is defined as $\pmb{\xi}=[x,y,z]^\mathsf{T}$, the attitude is defined in the roll-pitch-yaw convention as $\pmb{\eta}=[\phi,\theta,\psi]^\mathsf{T}$, and the angular velocity is defined as $\pmb{\nu} = [p,q,r]^\mathsf{T}$.
\subsection{Platform Dynamics Model}
 The dynamics model of this platform can be simplified as,
\begin{equation}
    \begin{bmatrix}
        \prescript{W}{}{\ddot{\pmb{\xi}}} \\  \prescript{B}{}{\dot{\pmb{\nu}}}
    \end{bmatrix}
    =
    \begin{bmatrix}
       {\frac{1}{m}}\prescript{W}{B}{\pmb{R}}&0\\
       0 & \prescript{B}{}{\pmb{I}}^{-1}
    \end{bmatrix}
    \pmb{u}
    +
    \begin{bmatrix}
        ^{W}\pmb{G}\\\pmb{0}
    \end{bmatrix},
    \label{eq: simplified dynamics}
\end{equation}
where $m$ is the total mass of the platform, 
$\pmb{G}$ is the gravitational acceleration,
$\prescript{W}{B}{\pmb{R}}$ is the rotation matrix from $\mathcal{F}_B$ to $\mathcal{F}_W$, $\pmb{I}$ is the inertia matrix of the platform. And 
\begin{equation}
    \pmb{u} = 
    \begin{bmatrix}
        \sum_{i=1}^{4} \prescript{B}{Q_i}{\pmb{R}}\,T_i \pmb{\hat{z}} \\
        \sum_{i=1}^{4} (\pmb{d}_i \times \prescript{B}{Q_i}{\pmb{R}}\,T_i \pmb{\hat{z}}) \\
    \end{bmatrix}
    =    
    \begin{bmatrix}
        \pmb{J}_\xi(\pmb{\alpha}) \\ \pmb{J}_\nu(\pmb{\alpha})
    \end{bmatrix}
    \pmb{T},
    \label{eq:bu_u}
\end{equation}   
where $\pmb{d}_i$ is the distance vector from $\mathcal{F}_B$'s center to $\mathcal{F}_{Q_i}$, 
and
\begin{equation}
\small
\begin{aligned}
      \pmb{J}_\xi&=
    \begin{bmatrix}
        -\sin{\alpha_{0}} &0 &\sin{\alpha_{2}} & 0 
        \\
        0 & \sin{\alpha_{1}} &0 & -\sin{\alpha_{3}} \\
        \cos{\alpha_{0}} &\cos{\alpha_{1}} & \cos{\alpha_{2}}& \cos{\alpha_{3}} 
    \end{bmatrix},\\  
    \pmb{J}_\nu&=l
    \begin{bmatrix}
    -\cos\alpha_0 & 0 & \cos\alpha_2 & 0\\
    0 & \cos\alpha_1 & 0 & -\cos\alpha_3\\
    \sin\alpha_0 & \sin\alpha_1 & \sin\alpha_2 & \sin\alpha_3
    \end{bmatrix},\\
    \pmb{\alpha}&=
    \begin{bmatrix}
    \alpha_0 & \alpha_1 & \alpha_2 & \alpha_3
    \end{bmatrix}^\mathsf{T}, \quad
    \pmb{T}=
    \begin{bmatrix}
    T_0 & T_1 & T_2 & T_3
    \end{bmatrix}^\mathsf{T},
\end{aligned}
\end{equation}
$l$ is the distance from the geometric center of each quadcopter to the geometric center of the central frame, 
$\alpha_i$ is the tilting angle of quadcopter $i$ (denoted as $\mathcal{Q}_i$), 
and $T_i$ is the magnitude of the thrust generated by $\mathcal{Q}_i$.  

\subsection{Actuator Dynamics}
For each $\mathcal{Q}_i$, the four rotating propellers collectively generate an independent force and torque output according to: 
\begin{equation} 
%\small
\label{eq: quad input decouple}
    \begin{bmatrix} 
        T_{i} \\ M_i^x \\ M_i^y \\ M_i^z
    \end{bmatrix} 
    =
    \begin{bmatrix} 
        1      &  1      &  1      & 1 \\
       -b      & -b      &  b      & b \\
        b      & -b      & -b      & b \\
       c_\tau & -c_\tau &  c_\tau & -c_\tau
    \end{bmatrix}
    \begin{bmatrix} 
        t_{i0} \\ t_{i1} \\ t_{i2} \\ t_{i3}
    \end{bmatrix},
\end{equation}
where $M^x_i$, $M^y_i$, and $M^z_i$ are the torque outputs in $\mathcal{F}_{Q_i}$; 
$b$ is a constant defined as $b = {a}/{\sqrt{2}}$ with $a$ the arm length of the quadcopter; 
$c_\tau$ is a constant defined as $c_\tau = {K_\tau}/{K_T}$ with ${K_\tau}$ the propeller drag constant and $K_T$ the propeller thrust constant; and 
$t_{ij}$ is the thrust force generated by propeller $j$ (denoted as $\mathcal{P}_j$) of $\mathcal{Q}_i$, 
defined by: $t_{ij} = K_T \omega_{ij}^2$,
where $\omega_{ij}$ is the rotational speed of $\mathcal{P}_j$ on $\mathcal{Q}_i$. The torque outputs $M_i^y$ are related to the hinge angles $\alpha_i$ through the tilting dynamics~\cite{pi2021simple}:
\begin{equation}
\label{eq: tilting dynamics}
    \Ddot{\alpha}_i = \frac{1}{I_i^y} M_i^y - \sin{\left(\frac{\pi}{2}i\right)}\dot{p} - \cos{\left(\frac{\pi}{2}i\right)}\dot{q},
\end{equation}
where $I_i^y$ is the inertia in the $\pmb{y}_{Q_i}$ direction.

\section{Nominal Control}
\label{sec:control}
\subsection{Hierarchical Architecture \& Tracking Control}
The overall controller features a hierarchical structure, consisting of
(i) a high-level controller that provides the desired wrench commands for the platform to track a reference trajectory, and maps these commands to the inputs of each thrust generator through control allocation;
(ii) a low-level controller for each quadcopter to achieve rapid response in tracking the desired joint angle and thrust.
The stability of this controller has been demonstrated in our previous work, please refer to~\cite{ruan2023control} for more details.

In high-level control, feedback linearization is implemented, and the six \ac{dof} wrench command is designed as follows: 
\begin{equation}
%\small
    \pmb{u}^{d}    
    =
    \begin{bmatrix}
        \pmb{J}_\xi \\ \pmb{J}_\nu
    \end{bmatrix}
    \pmb{T}
    =
    \begin{bmatrix}
        {m}\prescript{W}{B}{\pmb{R}}^\mathsf{T} & \pmb{0} \\ \pmb{0} & ^{B}\pmb{I}
    \end{bmatrix}    
    \left(\begin{bmatrix}
        \pmb{u}_{\xi} \\ \pmb{u}_{\nu}
    \end{bmatrix} -    
    \begin{bmatrix}
        \prescript{W}{}{\pmb{G}}\\\pmb{0}
    \end{bmatrix}\right),
    \label{eq: feedback}
\end{equation}
where the superscript $d$ indicates the desired values, $\pmb{u}_\xi$ and $\pmb{u}_\nu$ are virtual inputs for position control and attitude control, respectively. Combining \cref{eq: feedback} with \cref{eq: simplified dynamics}, the platform dynamics is equivalent to a double integrator and can be written in a state-space form as
\begin{equation} 
\label{eq:ss_lqi}
    \pmb{\dot{\mathcal{X}}} = \pmb{\mathcal{A}} \pmb{\mathcal{X}} + \pmb{\mathcal{B}}  \pmb{\widetilde{u}}, 
\end{equation}
where
\begin{equation}
\small
  \begin{aligned}
    \pmb{\mathcal{A}} &= 
    \begin{bmatrix}
        0 & 0 & \pmb{I}_3 & 0 \\
        0 & 0 & 0 & \pmb{I}_3 \\
        0 & 0 & 0 & 0 \\ 
        0 & 0 & 0 & 0
    \end{bmatrix}, \quad
    \pmb{\mathcal{B}} = 
    \begin{bmatrix}
        0 & 0 \\ 0 & 0 \\ \pmb{I}_3 & 0 \\ 0 & \pmb{I}_3
    \end{bmatrix}, \\
    \pmb{\mathcal{X}} &= 
    \begin{bmatrix} \pmb{\xi}^\mathsf{T}&\pmb{\eta}^\mathsf{T}&\pmb{\dot{\xi}}^\mathsf{T}&\pmb{\nu}^\mathsf{T}  
    \end{bmatrix}^\mathsf{T}, \quad
    \pmb{\widetilde{u}} =  
    \begin{bmatrix}
    \pmb{u}_\xi^\mathsf{T} & \pmb{u}_\nu^\mathsf{T}
    \end{bmatrix}^\mathsf{T}.
\end{aligned}  
\end{equation}
We design a LQI control scheme~\cite{yu2023compensating,kose2020simultaneous} to close the control loop with 
the augmented system states as
\begin{equation}
%\small
   \pmb{\widetilde{\mathcal{X}}}=\begin{bmatrix}
     \pmb{e}_\xi^{\mathsf{T}}& 
     \pmb{e}_\eta^{\mathsf{T}} &
     \dot{\pmb{e}}_\xi^{\mathsf{T}} &
     \dot{\pmb{e}}_\eta^{\mathsf{T}}  &
     \int\pmb{e}_\xi dt^{\mathsf{T}} &
     \int\pmb{e}_\eta dt^{\mathsf{T}}
    \end{bmatrix}^{\mathsf{T}},
\end{equation}
with
\begin{equation}
\begin{aligned}
    \pmb{e}_\xi=&\pmb{\xi}^r - \pmb{\xi},\\     \dot{\pmb{e}}_\xi= &\dot{\pmb{\xi}}^r - \dot{\pmb{\xi}},\\
    \pmb{e}_\eta=& \frac{1}{2}[\pmb{R}(\pmb{\eta})^T\pmb{R}(\pmb{\eta}^r)-\pmb{R}(\pmb{\eta}^r)^T\pmb{R}(\pmb{\eta})]_\vee, \\
    \dot{\pmb{e}}_{\eta}=& \pmb{R}(\pmb{\eta})^T\pmb{R}(\pmb{\eta}^r)\pmb{\nu}^r - \pmb{\nu}, \\
\end{aligned}
\end{equation}
where the superscript $r$ indicates the reference value, $\pmb{R}\left(\cdot\right)$ is the transformation from Euler angles to a standard rotation matrix, 
and $\left[\cdot\right]_\vee$ is the mapping from SO(3) to $\mathbb{R}^3$.

The cost function is
\begin{equation}
    \pmb{\widetilde{J}}(\pmb{\widetilde{\mathcal{X}}},\pmb{\widetilde{u}}) = \int_{0}^{\infty} (\pmb{\widetilde{\mathcal{X}}}^\mathsf{T} \pmb{\widetilde{\mathcal{Q}}} \pmb{\widetilde{\mathcal{X}}} + \pmb{\widetilde{u}}^\mathsf{T} \pmb{\widetilde{\mathcal{R}}} \pmb{\widetilde{u}}) dt,
\end{equation}
where $\pmb{\widetilde{\mathcal{Q}}}$ and $\pmb{\widetilde{\mathcal{R}}}$ are designed matrices that determine the closed-loop dynamics, and the optimal input $\pmb{\widetilde{u}}$ is given by
\begin{equation}
    \pmb{\widetilde{u}} = -\pmb{K}\pmb{\widetilde{\mathcal{X}}},
\end{equation}
where $\pmb{K}$ is the solution to the algebraic Riccati equation of the augmented system.

\subsection{Nominal Allocation Framework}
\subsubsection{Force Decomposition-based Allocation} \label{sec:FD allo}
Given $\pmb{u}^{d}$, we aim to determine the desired tilting angle $\pmb{\alpha}$ and thrust $\pmb{T}$ for the four quadcopters through the nonlinear mapping \cref{eq: feedback}\textemdash this process is known as the ``allocation problem''. 
One heuristic solution employs \ac{fd} to transform this nonlinear mapping problem into a linear one by defining intermediate variables~\cite{kamel2018voliro,yu2022over}: 
\begin{equation}
    \pmb{F}=
    \begin{bmatrix}
        F_{s0} & F_{c0} &\ldots & F_{s3} & F_{c3}
    \end{bmatrix}^\mathsf{T},
\end{equation}
where
\begin{equation}
    F_{si}=\sin\alpha_i\,T_i,\ 
    F_{ci}=\cos\alpha_i\,T_i.\
\end{equation}
With these new variables, \cref{eq: feedback} can be rewritten as
\begin{equation}
    \pmb{u}^{d}
    =
    \begin{bmatrix}
        \pmb{J}_\xi \\ \pmb{J}_\nu
    \end{bmatrix}\pmb{T}
    =
    \pmb{W}\pmb{F}
    \label{eq: allocation constraint},
\end{equation}
where $\pmb{W}\in\mathbb{R}^{6\times8}$ is a constant allocation matrix with full row rank. The general solution of $\pmb{F}$ is expressed as
\begin{equation}
\pmb{F}=\pmb{W}^\dagger\pmb{u}^d+\pmb{N}_W\pmb{Z},
    \label{eq: general solution}
\end{equation}
where $\pmb{N}_W\in\mathbb{R}^{8\times2}$ is the nullspace of $\pmb{W}$ and $\pmb{Z}\in\mathbb{R}^{2\times1}$ is an arbitrary vector. 
The least-squares solution can be acquired to minimize 
$\|F\|^2 = \|T\|^2$ by setting $\pmb{Z}=0$. 
The real inputs $T_i$ and $\alpha_i$ for 
the low-level controller can then be recovered as
\begin{equation}
        T_{i}=\sqrt{F_{si}^2+F_{ci}^2},\
    \alpha_i=\text{atan2}(F_{si},\,F_{ci}).
     \label{eq: determine alpha}
\end{equation}

\subsubsection{Thrust Force Saturation Issue}
\label{sec:saturation}
However, this \ac{fd}-based allocation framework (referred to as the nominal allocation framework) does not account for input constraints. Specifically, it could generate a desired thrust exceeding motor saturation, leading to platform instability. This issue, known as thrust force saturation, was investigated previously in~\cite{ruan2023control,ruan2020independent}, where the nominal \ac{fd}-based allocation framework proved inadequate for utilizing the full thrust capability of the platform, and an analytical solution was provided for a one-dimensional rotation scenario by formulating it as a min-max optimization problem. In this study, we generalize this problem for a standard trajectory-tracking scenario. At each timestep, the nullspace-based allocation framework (which will be introduced next) determines the optimal tilting angle and thrust for each quadcopter, subject to a predefined cost function and input constraints (see \cref{sec:saturation_exp}).

\subsection{Nullspace-based Allocation Framework}
In our previous work, 
we proposed a nullspace-based allocation framework~\cite{su2021nullspace} 
which has the advantages of both the \ac{fd}-based and \ac{qp}-based allocation frameworks while avoiding their known issues. 
In this framework, a QP problem is first formulated at each time step as: 
\begin{equation}
    \min_{{\Delta}\pmb{X},\pmb{s}} \quad
    \pmb{J}={\Delta}\pmb{X}^\mathsf{T}\pmb{P}{\Delta}\pmb{X}+\pmb{s}^\mathsf{T}\pmb{Q}\pmb{s}
    \label{eq: object function}
\end{equation}
\begin{equation}
%\small
\textrm{s.t.} \quad
    \pmb{W}\left( \pmb{s}+\pmb{F}(\pmb{\alpha}_o,\pmb{T}_o)+\left.\frac{\partial{\pmb{F}}}{\partial\pmb{X}}\right\vert_{\pmb{X}=\pmb{X}_o}
     \Delta\pmb{X}\right)=\pmb{u}^d
\label{eq:constraint}
\end{equation}
\begin{equation}
    \pmb{X}_{\text{min}}-\pmb{X}_o\leq {\Delta}\pmb{X} \leq \pmb{X}_{\text{max}}-\pmb{X}_o
    \label{eq: deltax1}  
\end{equation}
\begin{equation}
    {\Delta}\pmb{X}_{\text{min}}\leq {\Delta}\pmb{X} \leq {\Delta}\pmb{X}_{\text{max}}
    \label{eq: deltax2}  
\end{equation}
where \cref{eq: object function} is the object function, with $\pmb{X}$ defined as
\begin{equation}
  \pmb{X}=
  \begin{bmatrix}
    \pmb{\alpha}^\mathsf{T}&
    \pmb{T}^\mathsf{T} 
  \end{bmatrix}^\mathsf{T},  
\end{equation}
$\pmb{P}$ and $\pmb{Q}$ are weighting matrices.
\cref{eq:constraint} uses first-order linearization to approximate the nonlinear equality constraint \cref{eq: general solution}, where $\left[\cdot\right]_o$ is the value of a variable at previous timestep, 
$\Delta\left[\cdot\right]$ is the difference with respect to the previous timestep of a variable, and $\pmb{s}$ serves as a slack variable.
\cref{eq: deltax1,eq: deltax2} are two inequality constraints to limit the value of a variable or its rate of change.

The desired inputs for the current step can be approximated as,
\begin{equation}
    \pmb{X}=\pmb{X}_o+{\Delta}\pmb{X}.
    \label{eq: solve_alpha_T}
\end{equation}
Then, we can eliminate the approximation errors with the nullspace projection method,
\begin{equation}
    \pmb{F}^*= (\pmb{I}_{3n}-\pmb{N}_W\pmb{N}_W^\dagger)\pmb{W}^\dagger{\pmb{u}^d}+\pmb{N}_W\pmb{N}_W^\dagger\pmb{F}(\pmb{X}).
    \label{eq: proj}
\end{equation}
Finally, $\pmb{\alpha}^*$ and $\pmb{T}^*$ can be determined from $\pmb{F}^*$ using \cref{eq: determine alpha}. 
The nullspace-based allocation framework takes into account input constraints while still providing an exact solution for \cref{eq: allocation constraint}. 
This makes it more broadly widely than existing methods. For a detailed explanation of the implementation, the reader can refer to~\cite{su2021nullspace}. The \ac{ftc} utilizing this constrained allocation framework will be presented in \cref{sec:Robust control}.

\begin{figure*}[ht!]
    \centering
    \includegraphics[width=\textwidth]{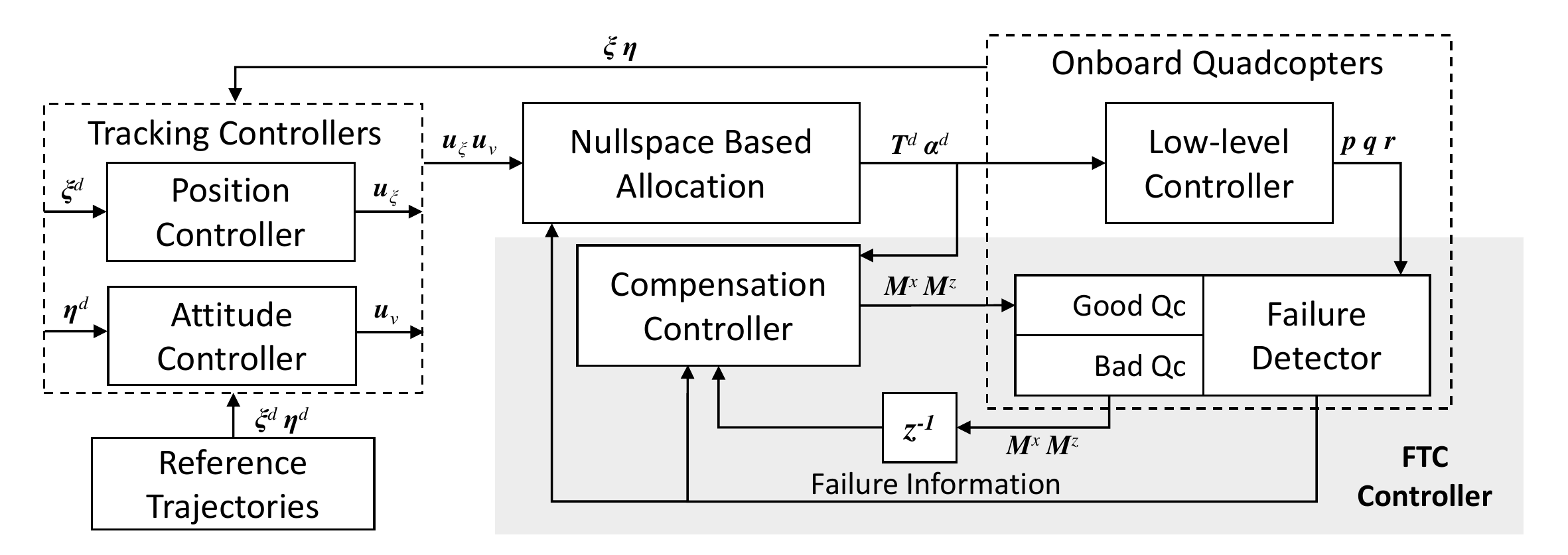}
    \caption{\textbf{Fault-Tolerant Controller Architecture.} Each quadcopter is equipped with onboard failure detection and motor control modules. When propeller failure is detected, the failure detection module communicates with the onboard controller to modify the low-level control strategy according to identified failure combinations. Additionally, it transmits failure information to the high-level controller, which leverages the nullspace-based control allocation to adjust thrust distribution among the four individual quadcopters.}
    \label{fig: control diagram}
\end{figure*}

\subsection{Low-level Control}
The thrust $T_i$ can be directly controllable in the low-level controller for each $\mathcal{Q}_i$, whereas the tilting angle $\alpha_i$ is controlled by $M_i^y$ through the motor's second-order rotational dynamics \cref{eq: tilting dynamics}. 
Therefore, a double-loop PID controller is applied to track the tilting angle~\cite{qian2020guidance,pi2021simple}:
\begin{equation} 
\label{eq:low_level_pid}
M_i^y = k_{D\alpha}\dot{e}_{\alpha} + k_{P\alpha}e_{\alpha} + k_{I\alpha}\int{e_{\alpha}}dt,
\end{equation}
where $k_{[\cdot]\alpha}$ are constant gains and the 
error term is defined as $e_{\alpha} = \alpha_{i}^d - \alpha_{i}$.

Considering these relationships and neglecting the fast motor dynamics that drive the propeller speeds,
the propeller thrusts can be calculated from \cref{eq: quad input decouple}:
\begin{equation}
%\small
    \begin{bmatrix}
    t_{i0} \\ t_{i1} \\ t_{i2} \\ t_{i3}
    \end{bmatrix}
    =
    sat\left(
    \begin{bmatrix}
    1  &  1  &  1 & 1 \\
    -b & -b &   b & b \\
     b & -b &  -b & b \\
    c_\tau & -c_\tau & c_\tau & -c_\tau
    \end{bmatrix}^{-1}
    \begin{bmatrix}
    T_{i} \\ M_i^x \\ M_i^y \\M_i^z
    \end{bmatrix}\right).
\label{eq:low_reverse}
\end{equation} 
where $sat(\cdot)$ is the saturation function. In the event of a failed propeller, the speed and thrust become nearly zero, subsequently altering the quadcopter's maximum thrust and moments.

\section{Fault-tolerant Controller} \label{sec:Robust control}
This section examines the \ac{ftc} control of the platform when some propellers are completely failed. 
Specifically, the focus lies on scenarios when one or two propellers on a single quadcopter have failed, while the other three quadcopters continue to function properly. 
Here, as an example, $\mathcal{Q}_3$ is assumed as the Bad QC. Of note, \ac{uav} fault detection has been extensively studied with various well-developed methods available in the literature~\cite{saied2015fault,zhang2021fault,lee2020fail}. Therefore, we assume that the propeller failure combinations have been accurately detected in this paper.  

\subsection{Fault-tolerant Controller Architecture}
As shown in \cref{fig: control diagram}, the \ac{ftc} controller comprises three primary components: (i) The low-level control on each quadcopter encompasses two onboard functions\textemdash one for propeller failure detection (assumed to be sufficiently fast), and another for quadcopter attitude and thrust control.
(ii) The compensation loop serves as an addition to the high-level position-attitude controller and employs auxiliary inputs to improve trajectory tracking performance and attenuate disturbances with a faster response within the saturation constraints. (iii) The nullspace-based control allocation framework is engaged to incorporate input constraints and adjust thrust force distribution among the four individual quadcopters.

When a propeller failure is detected, the low-level control module adjusts its strategy according to the failure combination to maintain control over thrust forces and tilting angle. Simultaneously, the high-level control allocation modifies thrust force distribution among the four individual quadcopters, considering their different thrust generation capabilities to prevent saturation. The compensation loop utilizes the auxiliary torque inputs of Good QCs to compensate for the disturbance torques generated by the low-level control of the Bad QC. 
% --------------------------------
\subsection{Propeller Failure Handling}

\subsubsection{One Propeller is Failed}
\label{sec:onefail}
\paragraph{\textbf{Low-level Control}}
In this case, three propellers on $\mathcal{Q}_3$ are assumed functional, while propeller $\mathcal{P}_0$ is considered to have failed. As a result, the four outputs in \cref{eq: quad input decouple} cannot be independently controlled~\cite{lee2020fail}. 
To account for this, \cref{eq:low_reverse} is adjusted to calculate the low-level commands as follows, where the control of $M_3^z$ is lost (its magnitude is relatively small),  
\begin{equation}
    \begin{bmatrix}
    t_{31}\\t_{32}\\t_{33}
    \end{bmatrix}
    =
    sat\left(\begin{bmatrix}
    \frac{1}{2} & -\frac{1}{2b}& 0\\[0.5em]
    0& \frac{1}{2b}& -\frac{1}{2b}\\[0.5em]
    \frac{1}{2}& 0& \frac{1}{2b}
    \end{bmatrix}
    \begin{bmatrix}
    T_{3}\\ M_3^x\\ M_3^y
    \end{bmatrix}\right). 
  \label{eq: lowel level 1}
\end{equation}
Given that the mapping matrix in \cref{eq: lowel level 1} is in full-rank, $T_3$, $M_3^x$, and $M_3^y$ can be controlled independently using the three remaining thrusts, without considering the saturation of $t_{3j}$.  However, in practice, the occurrence of $t_{32}\leq0$ as of result of $t_{32}=\frac{1}{2b}(M_3^x-M_3^y)$ is evidently unreasonable. Therefore, the control of the tilting angle with $M_3^y$ must take precedence over the toque $M_3^x$.

Based on the aforementioned analysis, \cref{eq: lowel level 1} is not employed in the low-level control. 
Instead, it is reformulated as follows:
\begin{equation}
    \begin{bmatrix}
    t_{31}\\t_{32}\\t_{33}
    \end{bmatrix}
    =  
    sat\left(
    \begin{bmatrix}
    \frac{1}{4} & -\frac{1}{4b}\\[0.5em]
    \frac{1}{4} & -\frac{1}{4b}\\[0.5em]
    \frac{1}{2} & \frac{1}{2b}
    \end{bmatrix}
    \begin{bmatrix}
    T_{3}\\ M_3^y
    \end{bmatrix}\right).
    \label{eq: lowel level 2}
\end{equation}
This implies the loss of control over both $M_3^x$ and $M_3^z$. Although this strategy introduces a torque disturbance to the central frame ($M_3^x\neq0$ and $M_3^z\neq0$), it ensures control of $\alpha_3$. 
The magnitude of the torque disturbances, which are proportional to $T_3$, can be reduced by adjusting thrust distribution across all quadcopters 
using the nullspace-allocation framework \cite{su2021compensation}. 
In addition, this torque disturbance will be compensated by the add-on compensation loop involving the other quadcopters, which will be introduced later. 
The comparison between using \cref{eq: lowel level 1} and \cref{eq: lowel level 2} in low-level control will be shown in \cref{sec:simulation}.

\paragraph{\textbf{High-level Control}}
From \cref{eq: lowel level 2}, it can be shown that $T_3$ will not be evenly distributed among the three remaining propellers, with $\mathcal{P}_3$ contributing half of the total required thrust. 
Therefore, the maximum value of $T_3$ must be changed from $4t_{\text{max}}$ to $2t_{\text{max}}$, 
where $t_{\text{max}}$ denotes the maximum thrust of a single propeller. 
In essence, half of the thrust-generation capability is lost despite only one propeller failing on a quadcopter. Therefore, 
the maximum thrust vector $T_{\text{max}}$  in \cref{eq: deltax1}, must be changed from 
$t_{\text{max}}\cdot\left[4,4,4,4\right]^\mathsf{T}$ to $t_{\text{max}}\cdot\left[4,4,4,2\right]^\mathsf{T}$. Note that the existing allocation strategies \cite{ryll2014novel,kamel2018voliro,zhao2020enhanced,santos2021fast} cannot accommodate input constraints, necessitating the use of the nullspace-based constrained allocation framework. 

\paragraph{\textbf{Fast Compensation Loop}}
We utilize the auxiliary inputs $M_i^x$ and $M_i^z$ from the three Good QCs to formulate a compensation loop, mitigating disturbances induced by the low-level control of $\mathcal{Q}_3$ \cite{su2021fast}. The platform's complete rotational dynamics derived in~\cite{ruan2023control}, and neglecting $^{B}\pmb{\nu} \times (^{B}\pmb{I} \,^{B}\pmb{\nu})$, can be expressed as: 
\begin{equation}
    ^{B}\Dot{\pmb{\nu}} = ^{B}\pmb{I}^{-1}\left(\pmb{J}_v\pmb{T}+\pmb{J}^x_{M}\pmb{M}^x+\pmb{J}^z_{M}\pmb{M}^z\right),
    \label{eq: rotaion2}
\end{equation}
where
\begin{equation}
\small
\begin{aligned}
    \pmb{J}^x_M&=
    \begin{bmatrix}
      -\cos{\alpha_0} &0 & \cos{\alpha_2}&0 \\
      0 & \cos{\alpha_1} & 0 & -\cos{\alpha_3}\\
      \sin{\alpha_0}&\sin{\alpha_1}&\sin{\alpha_2}&\sin{\alpha_3}
    \end{bmatrix},\\
    \pmb{M}^x&=
    \begin{bmatrix}
    M_0^x & M_1^x & M_2^x & M_3^x
    \end{bmatrix}^\mathsf{T},\\
    \pmb{J}_M^z&=
    \begin{bmatrix}
      \sin{\alpha_0} &0 & -\sin{\alpha_2}&0 \\
      0 & -\sin{\alpha_1} & 0 & \sin{\alpha_3}\\
      \cos{\alpha_0}&\cos{\alpha_1}&\cos{\alpha_2}&\cos{\alpha_3}
    \end{bmatrix},\\    
    \pmb{M}^z&=
    \begin{bmatrix}
    M_0^z & M_1^z & M_2^z & M_3^z
    \end{bmatrix}^\mathsf{T}.
\end{aligned} 
\end{equation}

When a propeller fails, a \ac{qp} problem is formulated to solve for the optimal 
auxiliary inputs of $\mathcal{Q}_{0-2}$ for disturbance compensation. 
The equality constraint is designed as
\begin{equation}
    \pmb{J}^x_M\pmb{M}^x+\pmb{J}^z_M\pmb{M}^z+\pmb{k}=0,
\end{equation} 
where $\pmb{k}$ is a slack variable. The object function is
\begin{equation}
    \pmb{J}(\pmb{y},\pmb{k})
     = \pmb{y}^\mathsf{T}\pmb{A}\pmb{y}+\pmb{k}^\mathsf{T}\pmb{B}\pmb{k},
\end{equation}
where $\pmb{y}$ is defined as
\begin{equation}
    \pmb{y}=
    \begin{bmatrix}
    M_0^x & M_1^x & M_2^x & M_0^z & M_1^z & M_2^z
    \end{bmatrix}^\mathsf{T},
\end{equation}
and $\pmb{A}$ and $\pmb{B}$ are constant, 
positive semi-definite gain matrices. 
Saturation is included as the following inequality constraints:
\begin{equation}
  \resizebox{0.88\linewidth}{!}{%
     $0\leq
    \begin{bmatrix}
    1  &  1  &  1 & 1 \\
    -b & -b &   b & b \\
     b & -b &  -b & b \\
    c_\tau & -c_\tau & c_\tau & -c_\tau
    \end{bmatrix}^{-1}   
    \begin{bmatrix}
    T_{i}\\M_i^x \\M_i^y \\M_i^z
    \end{bmatrix}
    \leq t_\text{max}, \forall i=0, \,1,\,2.$
    }
\end{equation}
Note that for this QP problem, 
$M_3^x$ and $M_3^z$ from $\mathcal{Q}_3$ are used as feedback 
along with 
$T_i$ and $M_i^y$ from $\mathcal{Q}_{0-2}$.

Compared to other disturbance attenuation approaches for over-actuated \ac{uav} platforms~\cite{yang2021learning,wang2020practical}, which primarily compensate for unmodeled dynamics by adjusting the virtual wrench command via a second-order PID loop, this 
auxiliary-input-based compensation loop can be regarded as feed-through dynamics. Consequently, it enables a faster response to torque disturbances introduced by the low-level adjustment of Bad QC.   

\subsubsection{Two Propellers are Failed}
\paragraph{\textbf{Low-level Control}}
In this case, it is assumed that two propellers on $\mathcal{Q}_3$ have failed. In order to maintain control over $T_3$ and $M_3^y$, specific requirements must be imposed on the propeller failure combinations to ensure $M_3^y$ remains controllable. This framework is unable to address two situations that may occur when propeller failure affects both $\mathcal{P}_0$ and $\mathcal{P}_3$ or 
both $\mathcal{P}_1$ and $\mathcal{P}_2$ (\cref{tab:combination}).

As an example, we consider the case where both $\mathcal{P}_0$ and $\mathcal{P}_1$ 
are failed. The low-level controller is designed as 
\begin{equation}
    \begin{bmatrix}
    t_{32}\\t_{33}
    \end{bmatrix}
    =
    sat\left(
    \begin{bmatrix}
        1& 1\\
        -b& b
    \end{bmatrix}^{-1}
    \begin{bmatrix}
        T_{3}\\ M_3^y
    \end{bmatrix}\right).
    \label{eq: lowel level 3}
\end{equation}
In this case, $M_3^x$ and $M_3^z$ will be transferred to the central frame as disturbance torques with larger magnitudes than those in case of a single propeller failure. 
As in \cref{sec:onefail}, 
the high-level maximum thrust constraint is modified, and the compensation loop is implemented for disturbance attenuation.

Obviously, in the event of three or four propeller failures, the quadcopter effectively loses all control. Nonetheless, in cases of quadcopter-level failure (marked with $\times$ in \cref{tab:combination}), the platform remains controllable if the other Good QCs retain the controllability for the \ac{dof}s to be controlled, which must include at least the gravity direction~\cite{lee2020fail}.  For the four-quadcopter configuration investigated in this paper, the quadcopter opposing the failed one would also need to be disabled, and platform control would rely on the remaining pair of quadcopters. Specifically, the high-level control's thrust saturation limit for each quadcopter must be adjusted accordingly, with the thrust limits of the lost pair of quadcopters set to zero.  Finally, when two non-opposing quadcopters lost control due to the aforementioned propeller failures cases, the platform would fail.  Thus, all possible propeller failure scenarios have been addressed.

\begin{table}[t!]
\small
\caption{\textbf{Controllability of $M_3^y$ under Different Propeller Failure Combinations} ($M_3^y$ uncontrollable cases lead to quadcopter-level failure.)}
\centering
\resizebox{0.85\linewidth}{!}{
\begin{tabular}{c c c}
\toprule
\textbf{Group} &\textbf{Failure  Combination}& \textbf{$M_3^y$ Controllable?}\\
\midrule 
\multirow{1}{*}{One}
& & \checkmark\\
\midrule
\multirow{6}{*}{Two}
&0, 1& \checkmark\\
&0, 2& \checkmark\\
&0, 3& $\times$\\
&1, 2& $\times$\\
&1, 3& \checkmark\\
&2, 3& \checkmark\\
\midrule
\multirow{1}{*}{Three or Four}
& & $\times$ \\
\bottomrule
\end{tabular}}
\label{tab:combination}
\end{table}

\section{Simulation} 
\label{sec:simulation}

\subsection{Simulation Setup}
To compare the two low-level control adjustment strategies (\cref{sec:onefail}), 
a dynamic simulation was constructed using Matlab/Simulink. 
The Simscape Multibody module was employed to simulate the platform's complete dynamics. 
All known hardware characteristics were included in the simulated model, 
such as sampling frequencies, measurement noise, communication delay, motor dynamics, and more, with the full list provided in \cref{tab:setup}. Here, $m_0$ and $I_0$ refer to the mass and inertia matrix of the mainframe, while $m_i$ and $I_i$ refer to the mass and inertia matrix of each quadcopter with the passive hinge.

\begin{table}[ht!] 
\small
\centering
\caption{Physical and Software Properties in Simulation} \label{tab:setup} 
\begin{tabular}{cc}
\toprule
\textbf{Parameter} & \textbf{Value} \\ 
\midrule
$m_0$ & $0.036/kg$ \\
$m_i$ & $0.027/kg$ \\
$I_0$ & $\text{diag}\left(\left[3\ 3 \ 4.5\right]\right)/$kg\textperiodcentered cm\textsuperscript{2} \\
$I_i$ & $\text{diag}\left(\left[0.16\ 0.16\ 0.29\right]\right)/$kg\textperiodcentered cm\textsuperscript{2}  \\
$l$ & $0.14/m$ \\
$t_\text{max}$ & $0.167/N$ \\
communication delay & $0.02/s$ \\
\bottomrule
\end{tabular}
\end{table}

\begin{figure}[ht!]    
\centering
    \begin{subfigure}[b]{0.49\linewidth}
    \centering
        \includegraphics[width=\linewidth,trim=4cm 0.1cm 4.3cm 0cm, clip]{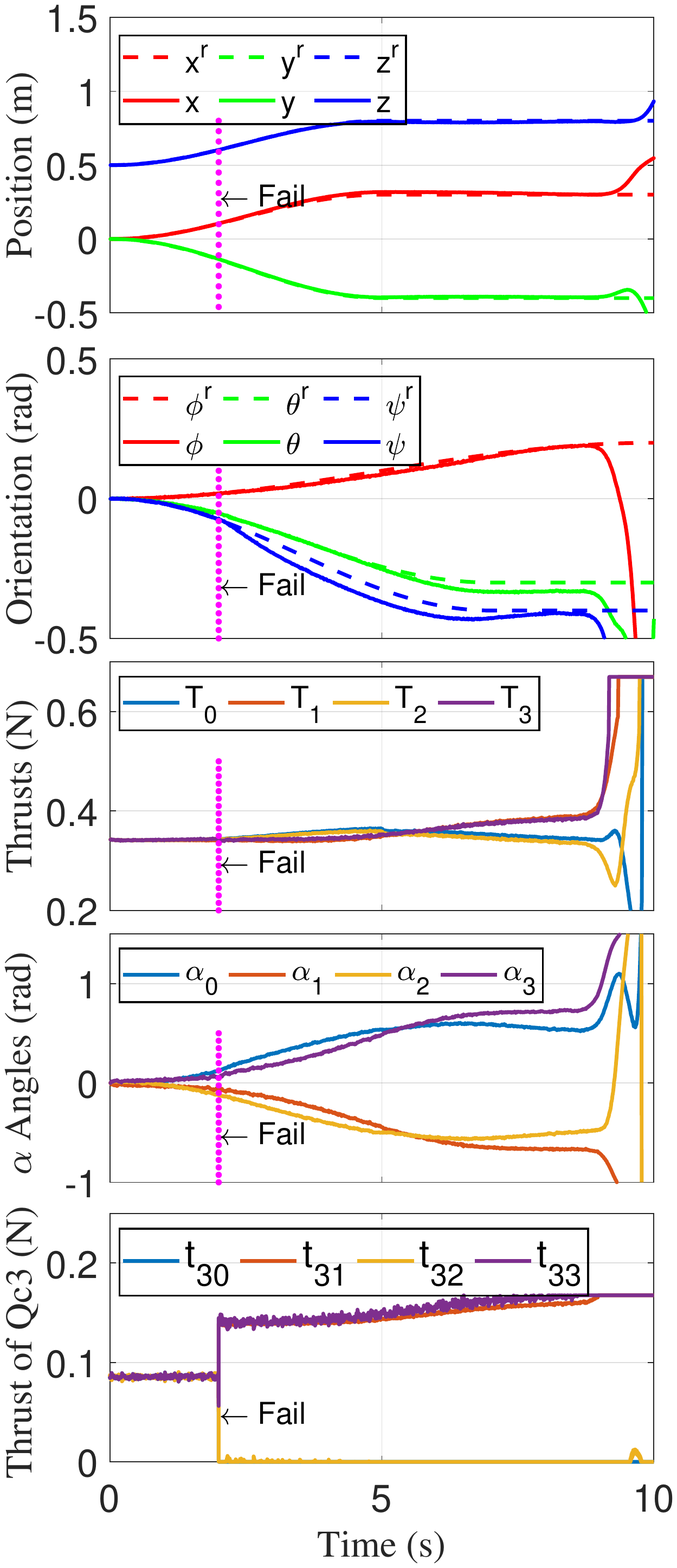}
        \caption{The first strategy.}
        \label{fig:sim_1_hinge_normal_xyz}
    \end{subfigure}
    \begin{subfigure}[b]{0.49\linewidth}
    \centering
        \includegraphics[width=\linewidth,trim=4cm 0.1cm 4.3cm 0cm, clip]{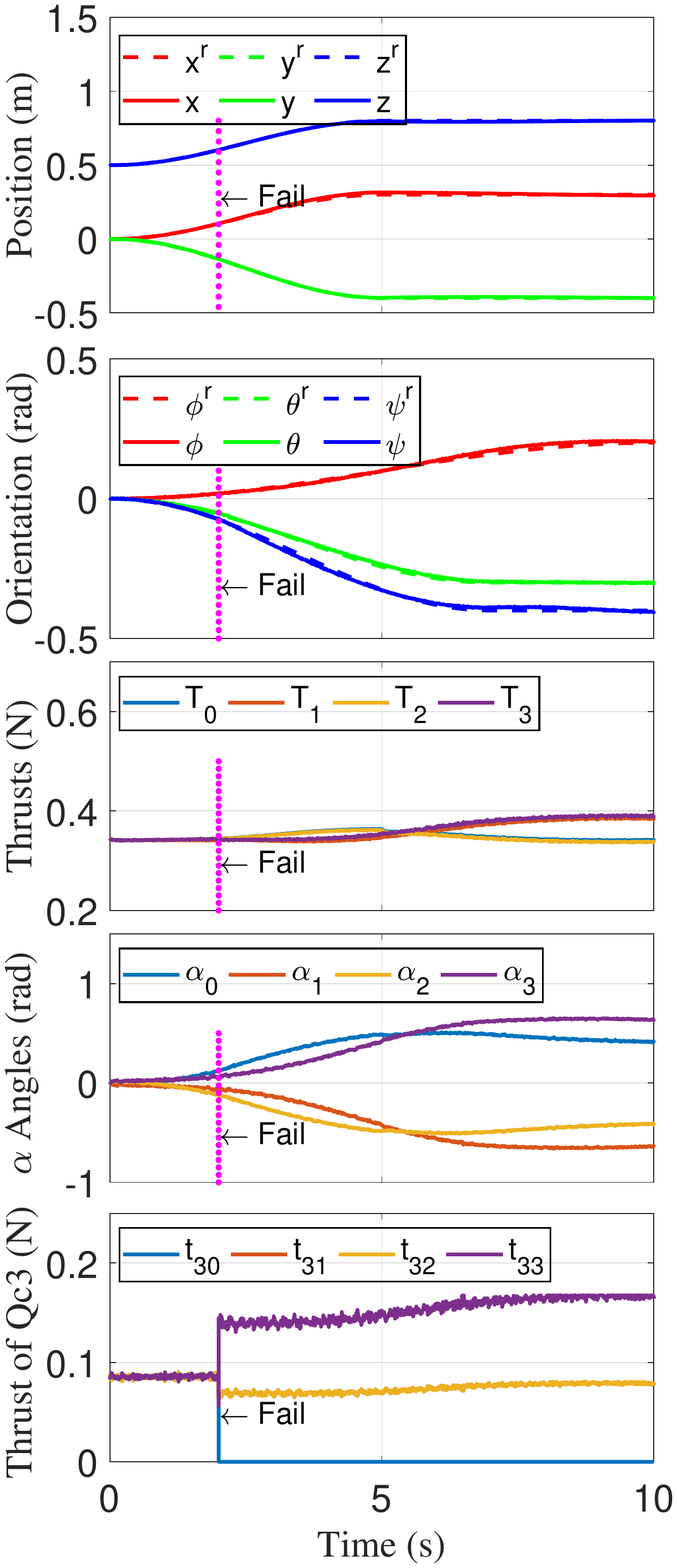}
        \caption{The second strategy}
        \label{fig:sim_2_hinge_normal_xyz}
    \end{subfigure}\\
    \caption{\textbf{Simulation: Trajectory tracking performance of two low-level adjustment strategies under propeller failure.} (a) corresponds to \cref{eq: lowel level 1} and (b) corresponds to \cref{eq: lowel level 2}.}
    \label{fig:sim_hinge}
\end{figure}

\subsection{Simulation Results}
Two low-level adjustment strategies (\cref{eq: lowel level 1} and \cref{eq: lowel level 2}) were compared in this simulation. The nominal controller with \ac{fd}-based allocation was utilized at the high-level to track the reference position and attitude trajectory. In both tests, $\mathcal{P}_0$ on $\mathcal{Q}_3$ began to fail at $2s$ (\cref{fig:sim_hinge}).

As illustrated in \cref{fig:sim_1_hinge_normal_xyz}, with the first strategy (\cref{eq: lowel level 1}), the desired thrust $t_{32}$ could become a non-positive value, which was subsequently set to zero due to propeller limitation. This deteriorated the regulation of the tilting angle. 
As a result, the high-level thrust finally saturated at approximately $9s$, leading to instability in the
position and attitude control of the whole platform. In contrast, with the second strategy (\cref{eq: lowel level 2}), both position and attitude control remained stable throughout the trajectory, as shown in \cref{fig:sim_2_hinge_normal_xyz}. 
In the low-level control of $\mathcal{Q}_3$, all propeller thrusts stayed within the saturation range, thus ensuring proper tilting angle control.

\section{Experiments} \label{sec:experiment}

\subsection{Experiment Setup} \label{sec: exp_setup}
The Crazyflie 2.1 quadcopter served as the basis for each individual unit within the platform. To achieve greater thrust force, the battery and motors on each quadcopter were upgraded, resulting in a maximum thrust force of $0.67~N (4{\times}t_\textit{max})$ and a mass (including the passive hinge) of $27~g$. 
The entire platform has a total mass of $144~g$, with overall dimensions of $36\times36\times6~cm$. 
A light-weighted tether was attached from the ceiling of the indoor environment to the center of the platform to protect the hardware in the case of failure. This tether remained loose and exerted negligible force and torque on the platform throughout all the experiments.

\begin{figure}[t!]
    \centering
    \includegraphics[width=\linewidth]{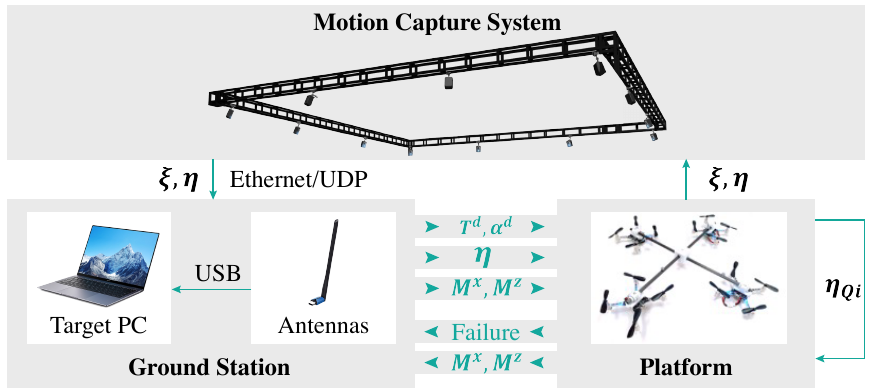}
    \caption{\textbf{Experimental setup.}
    A host PC runs the high-level controller at $100~Hz$ with platform position and attitude feedback from the OptiTrack motion-capture system. The onboard low-level controller runs at $500~Hz$ on each quadcopter.}
    \label{fig:system_diagram}
\end{figure}

An OptiTrack motion-capture system served as the external sensor for measuring the platform's position and attitude. 
The main controller operated on a ground-based computer and communicated through Ethernet with the motion-capture system 
to calculate $T_{i}$, $\alpha_{i}$, $M_i^x$, $M_i^z$ for each quadcopter. The radio-communication antenna transmitted these values, along with the attitude of the central frame, to the individual quadcopter. Every quadcopter was equipped with its own microprocessor, IMU, and modules for failure detection and onboard control. The system presumed the propeller failure combinations to be identifiable by the efficient failure detection module; it utilized the control module to adjust the low-level control strategy based on the propeller-failure combination and to regulate $T_{i}$ and $\alpha_{i}$. The experimental setup is shown in \cref{fig:system_diagram}.

\begin{figure}[t!]    
    \centering
    \begin{subfigure}[b]{0.49\linewidth}
    \centering
        \includegraphics[width=\linewidth,trim=4cm 4.8cm 4.2cm 0cm, clip]{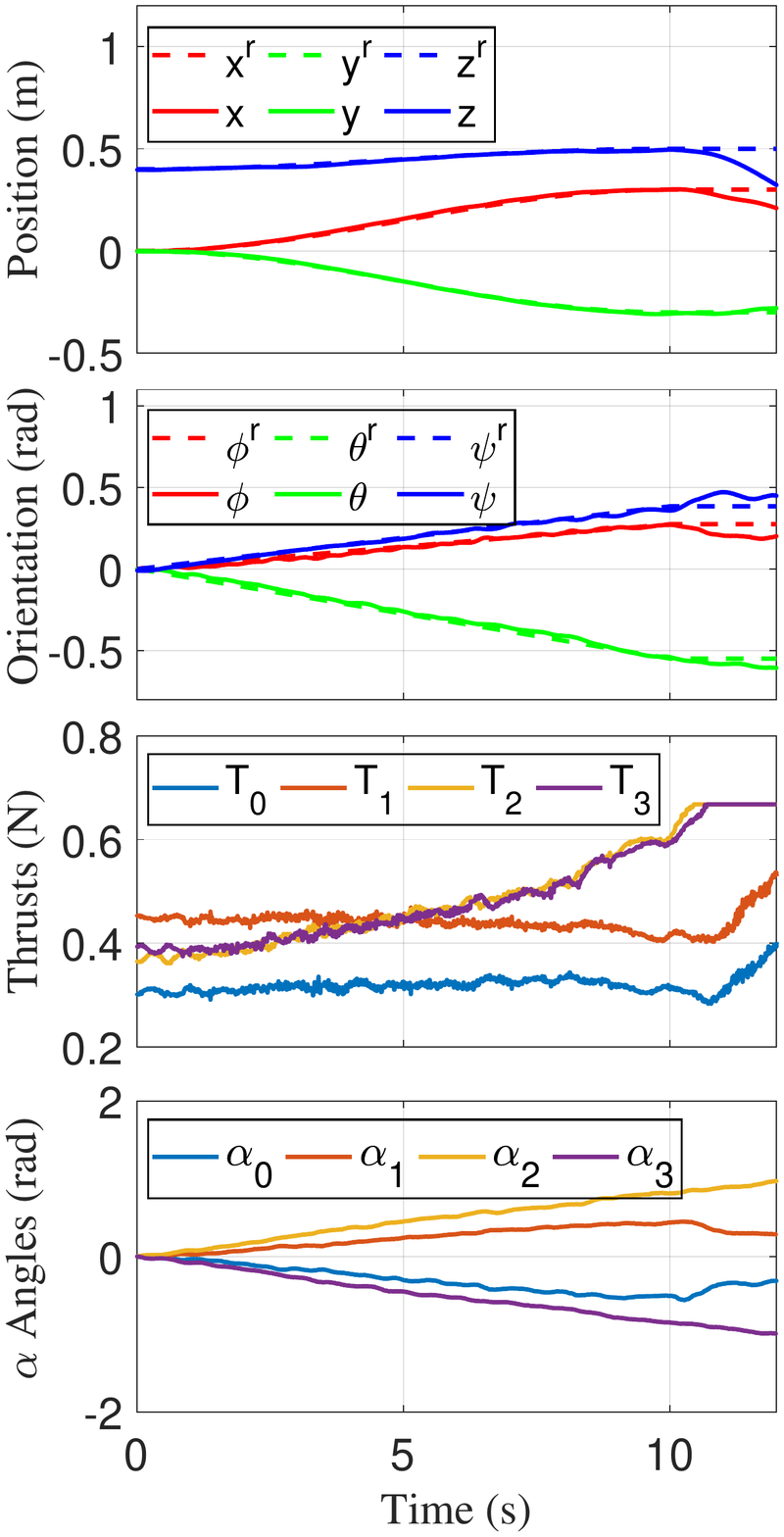}
        \caption{\ac{fd}-based framework.}
        \label{fig:hinge_normal}
    \end{subfigure}
    \begin{subfigure}[b]{0.49\linewidth}
    \centering
        \includegraphics[width=\linewidth,trim=4cm 4.8cm 4.2cm 0cm, clip]{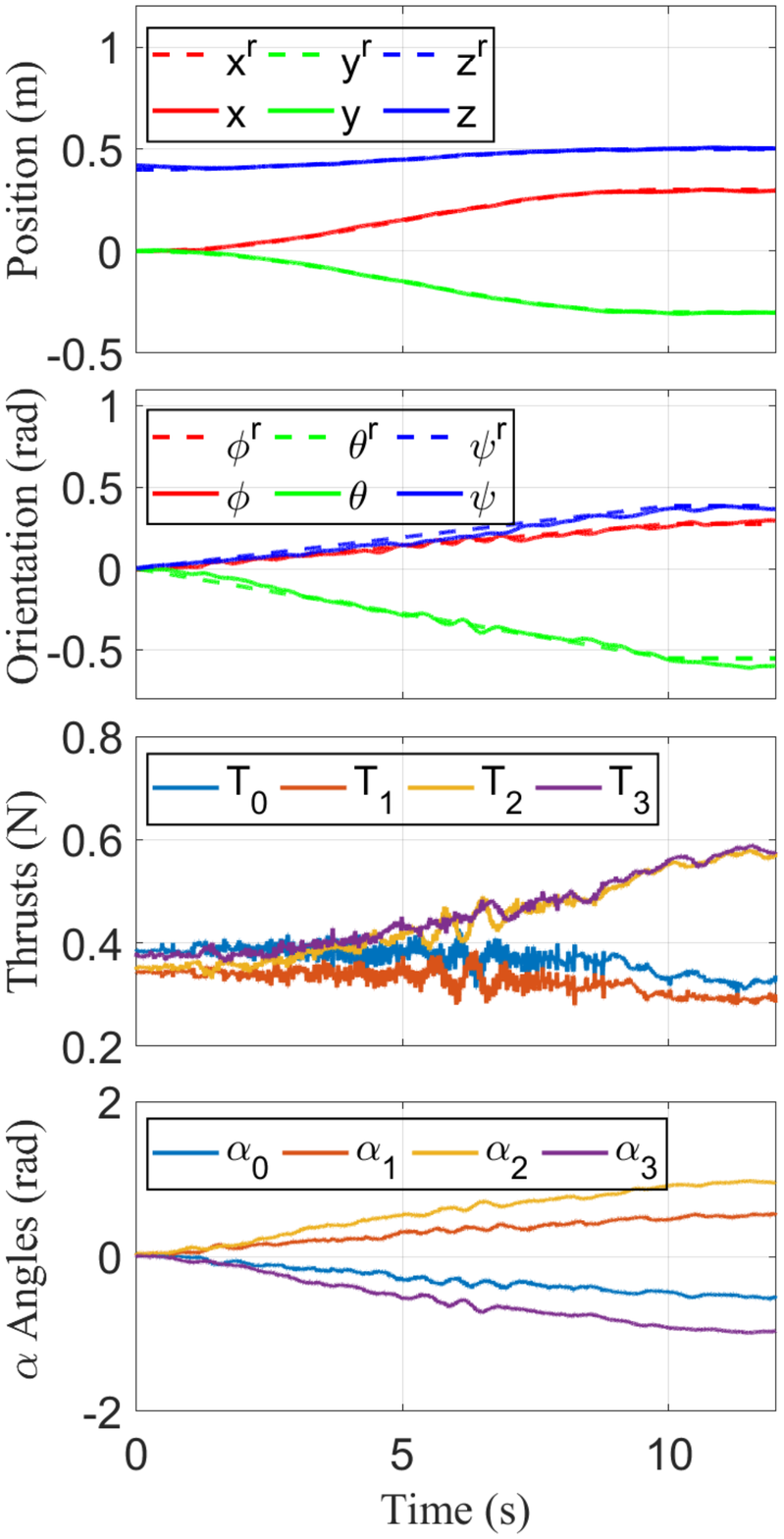}
        \caption{Nullspace-based framework.}
        \label{fig:hinge_opt}
    \end{subfigure}
    \caption{\textbf{Experiment: Using different allocation frameworks to handle thrust force saturation.}}
    \label{fig:hinge_experiment}
\end{figure}
\subsection{Thrust Force Saturation}
\label{sec:saturation_exp}
In this experiment (\cref{fig:hinge_experiment}), both \ac{fd}-based and nullspace-based allocation frameworks were implemented on the UAV platform to track a six \ac{dof} reference trajectory and remain at the final attitude for $2s$. As shown in \cref{fig:hinge_normal}, the \ac{fd}-based allocation framework could not address the thrust force saturation issue. Specifically, the desired thrust commands $T_2$ and $T_3$ continued to increase and saturated at approximately $10.7s$, 
leading to an unstable system. 
Meanwhile, $T_0$ and $T_1$ remained below $0.4~N$, 
implying that the platform could generate sufficient thrust to remain airborne. For the nullspace-based allocation framework, the platform successfully reached the desired attitude and maintained stability (\cref{fig:hinge_opt}). 
As analyzed in \cref{sec:saturation}, this experiment further demonstrates that the \ac{fd}-based framework does not fully exploit the platform's capabilities, while the nullspace-based framework enhances performance by incorporating the constraints.
\begin{figure}[t!]    
    \centering
    \begin{subfigure}[b]{0.49\linewidth}
    \centering
        \includegraphics[width=\linewidth,trim=3.8cm 5.5cm 3.5cm 1.5cm, clip]{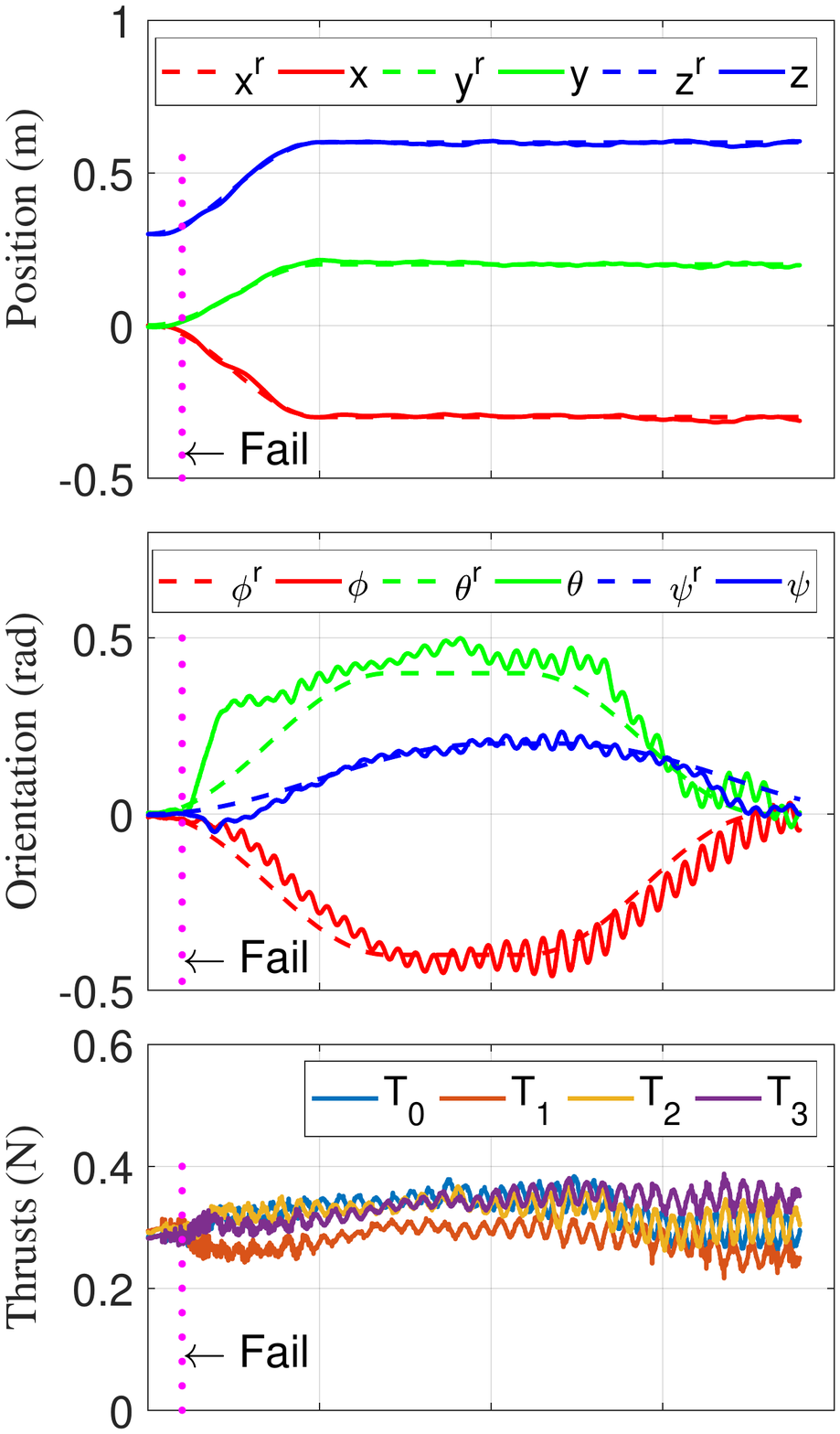}\\
        \includegraphics[width=\linewidth,trim=3.8cm 5.4cm 3.5cm 2.5cm, clip]{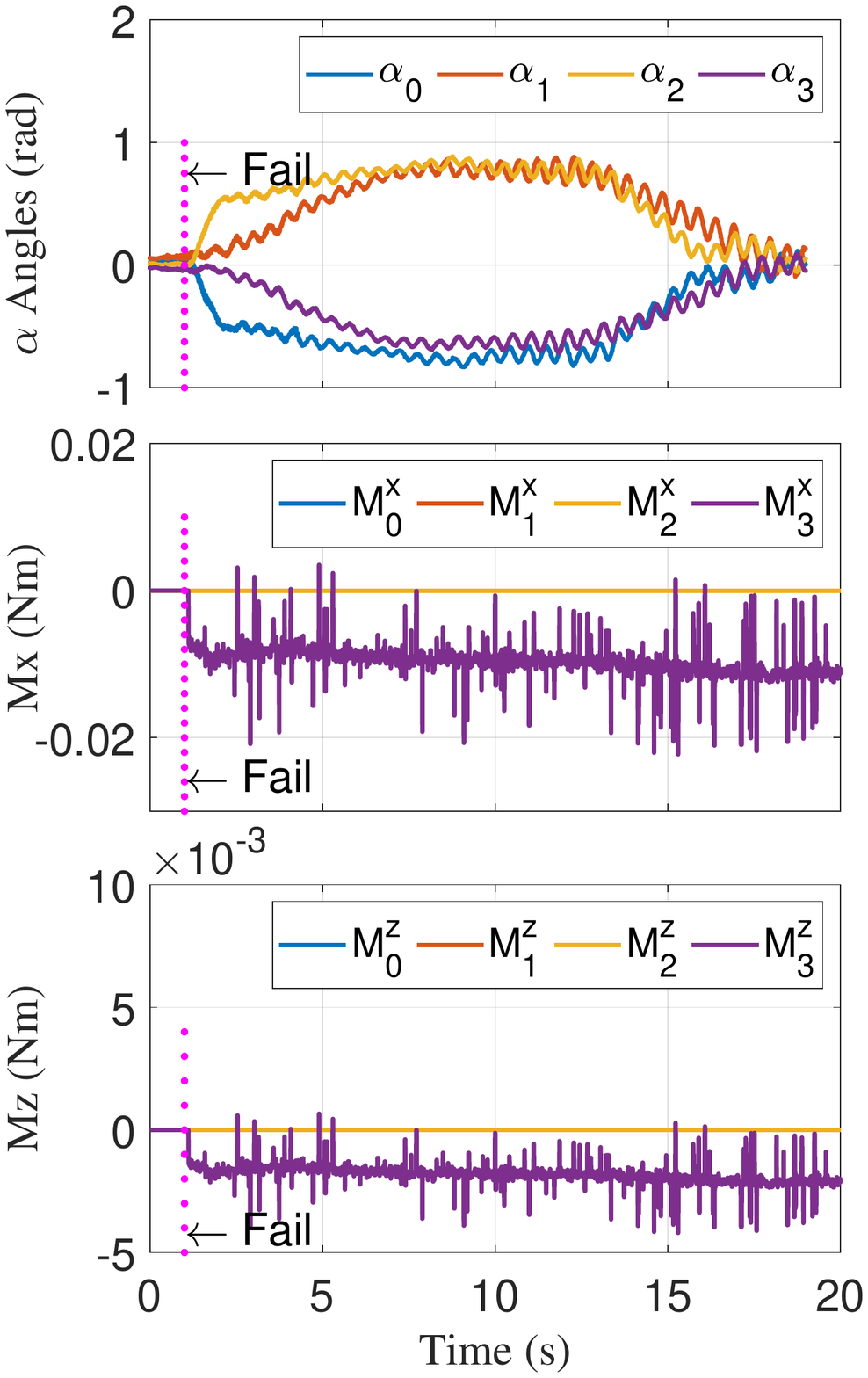}
        \caption{One Fail (U): N+L}
        \label{fig:hinge_normal_1fail}
    \end{subfigure}
    \begin{subfigure}[b]{0.49\linewidth}
    \centering
        \includegraphics[width=\linewidth,trim=3.8cm 5.5cm 3.5cm 1.5cm, clip]{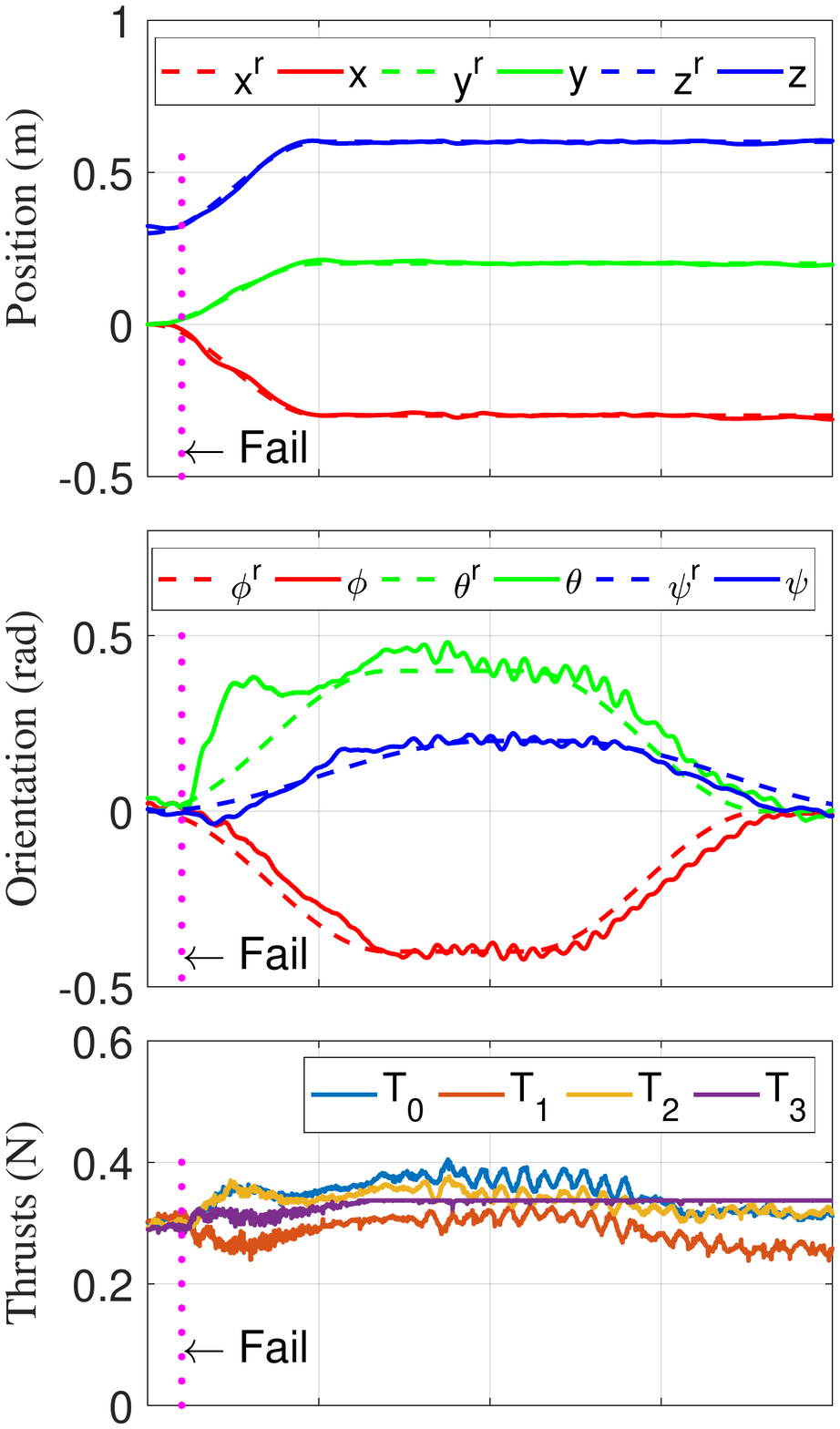}\\
        \includegraphics[width=\linewidth,trim=3.8cm 5.4cm 3.5cm 2.5cm, clip]{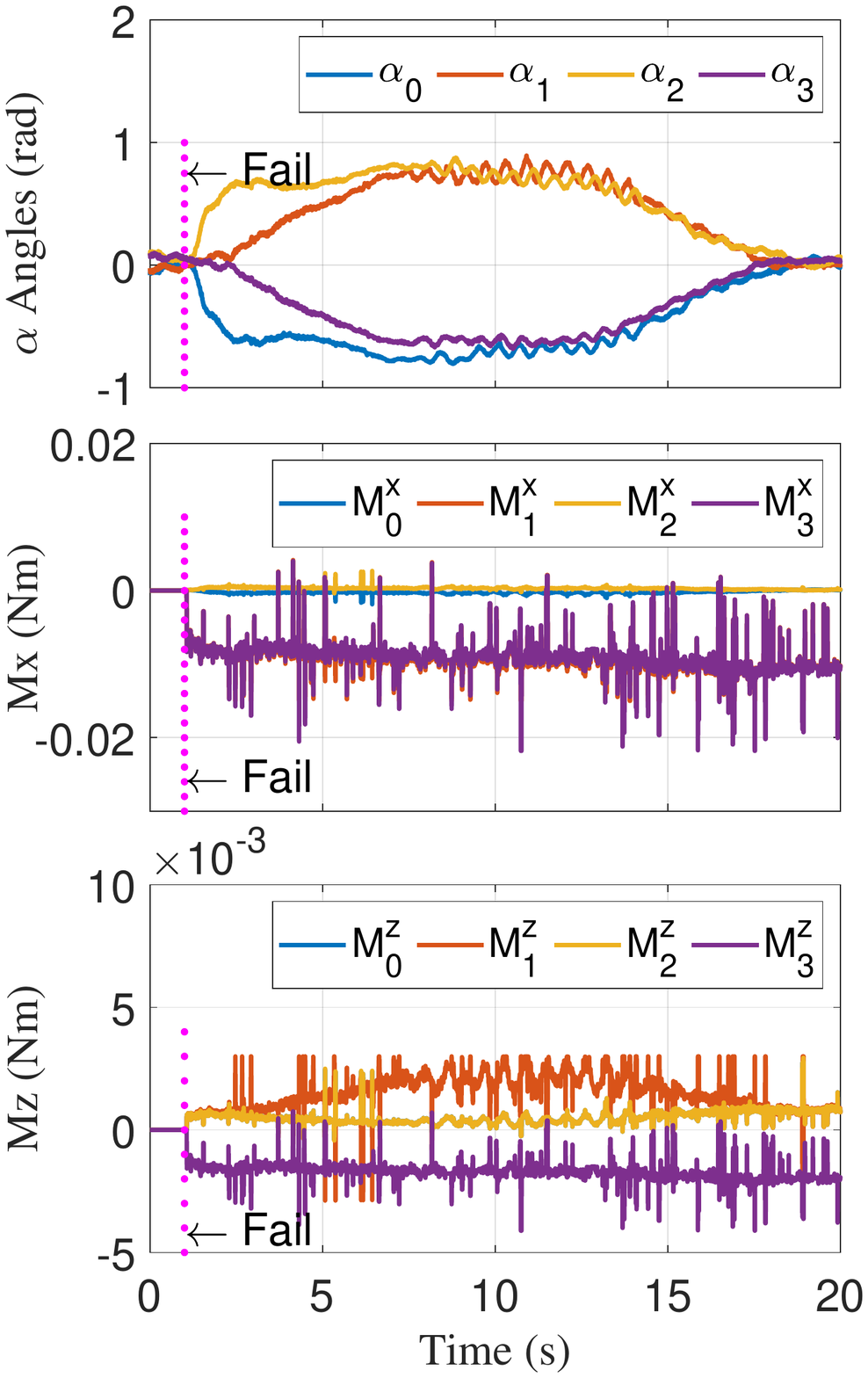}
        \caption{One Fail (U): FTC}
        \label{fig:hinge_opt_1fail}
    \end{subfigure}
    \caption{\textbf{Case 1: Trajectory tracking performance when one propeller is failed.} (U stands for the unsaturated trajectory that satisfies $T_3\leq2\,t_\text{max}$ all the time, N+L stands for the nominal controller + low-level adjustment, and FTC stands for the FTC framework we proposed.
    The same notations are applied for the rest of this paper.)}
    \label{fig:hinge_1fail}
\end{figure}
% --------------------------------
\subsection{Propeller Failure}
Three cases were designed to verify the effectiveness of the proposed \ac{ftc} controller in scenarios where one or two propellers failed during trajectory tracking. 
For each case, the nominal allocation strategy with low-level adjustment (denoted as N+L controller) is compared with the \ac{ftc} controller. Propeller failure was simulated by setting the speed of the corresponding propellers to zero.

\subsubsection{One Failed Propeller}
\paragraph{\textbf{Unsaturated Trajectory}}
In this experiment, 
an unsaturated trajectory was employed, ensuring that $T_3\leq2\,t_\text{max}$ at each timestep. 
As analyzed in \cref{sec:onefail}, the N+L controller maintained the stability of the platform when the trajectory did not trigger low-level saturation of $\mathcal{Q}_3$. 
In this situation, the main difference between the N+L controller and the \ac{ftc} controller was the fast compensation loop for disturbance rejection. The $\mathcal{P}_0$ on $\mathcal{Q}_3$ began to fail at $1s$. The tracking performance of the two controllers is plotted in \cref{fig:hinge_1fail}. 

\begin{figure}[t!]   
    \centering
    \begin{subfigure}[b]{0.49\linewidth}
    \centering
        \includegraphics[width=\linewidth,trim=3.8cm 3.6cm 3.5cm 1.5cm, clip]{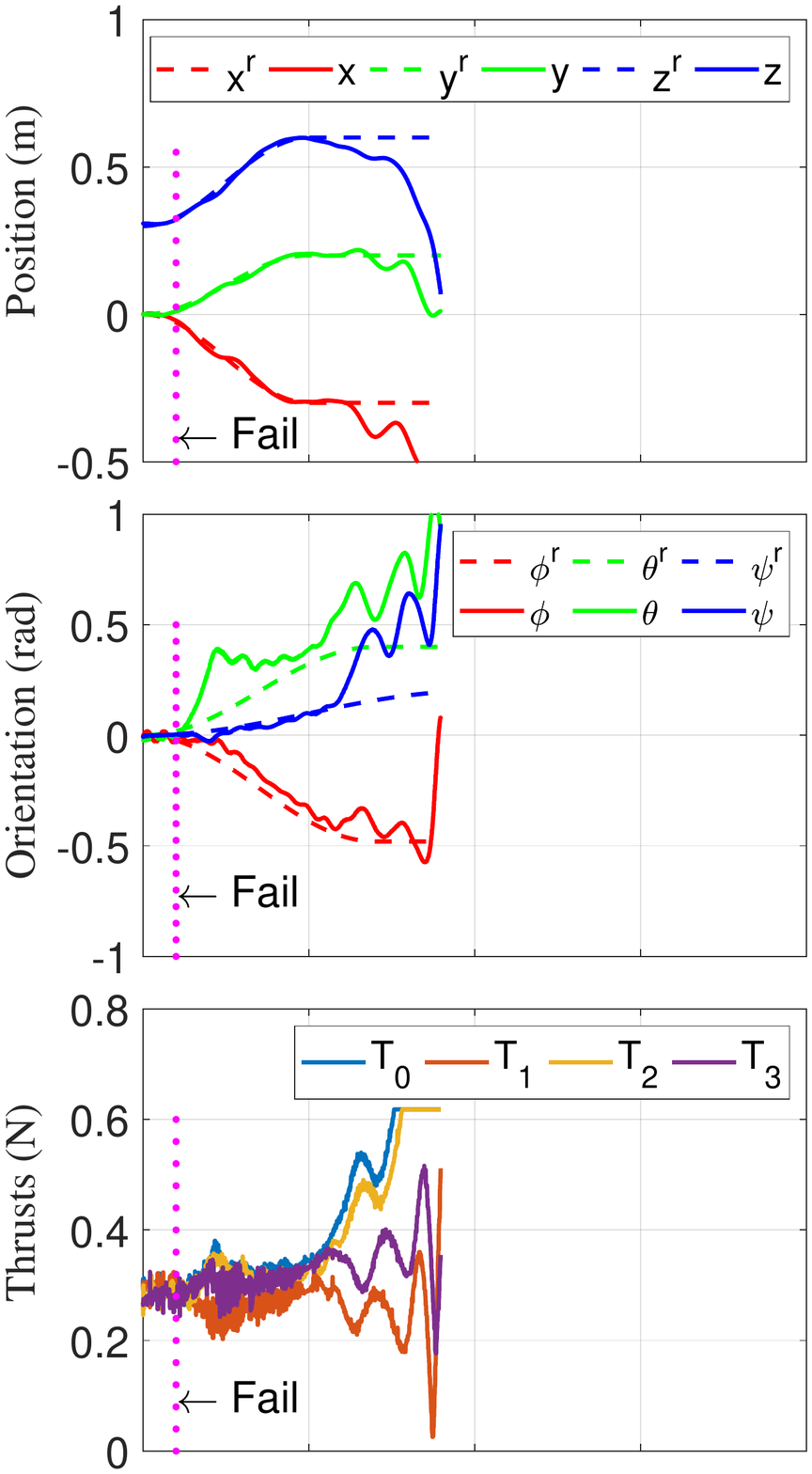}\\
        \includegraphics[width=\linewidth,trim=3.8cm 5.4cm 3.5cm 1.5cm, clip]{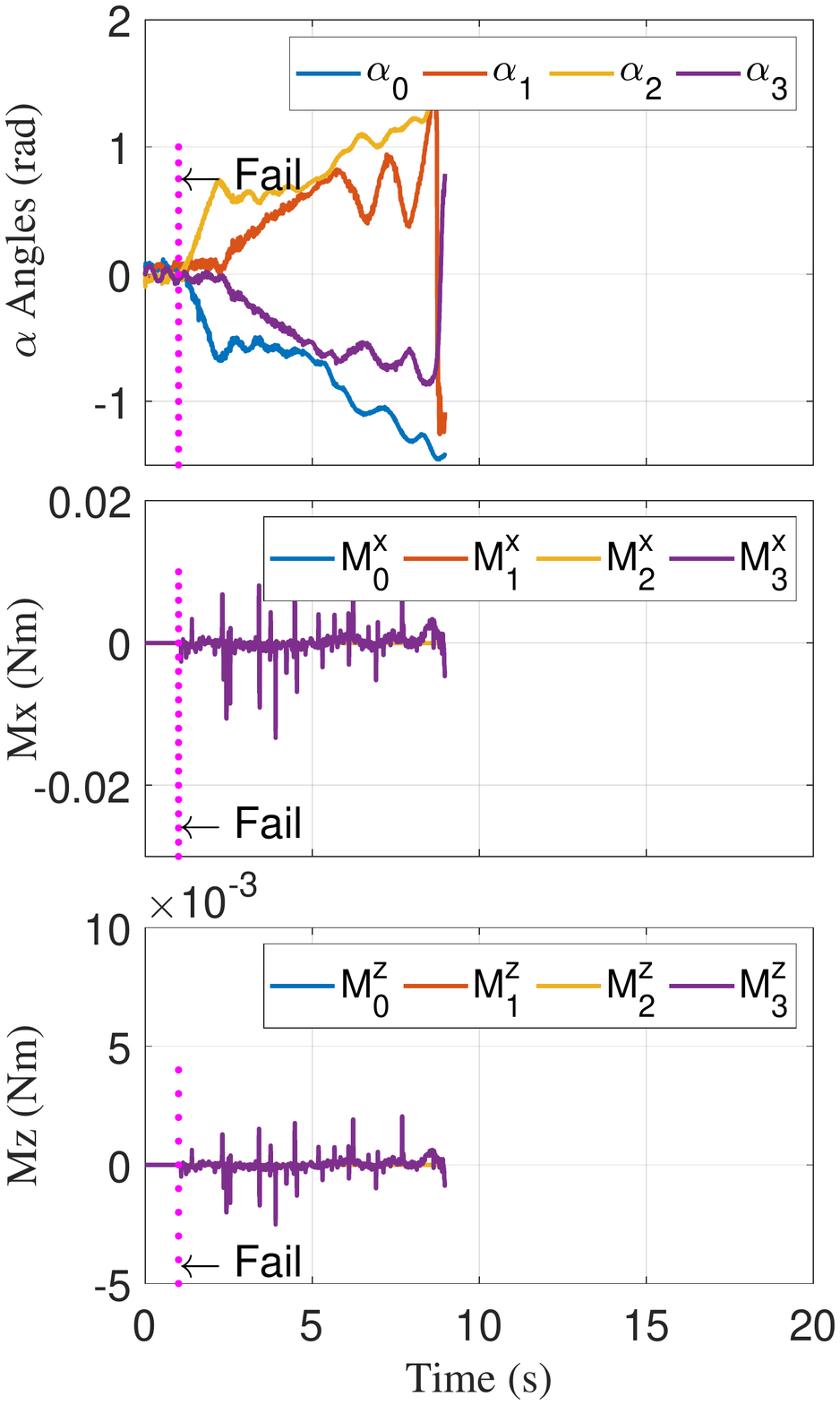}
        \caption{One Fail (S): N+L}
        \label{fig:hinge_normal_1fail_2}
    \end{subfigure}
    \begin{subfigure}[b]{0.49\linewidth}
    \centering
        \includegraphics[width=\linewidth,trim=3.8cm 3.6cm 3.5cm 1.5cm, clip]{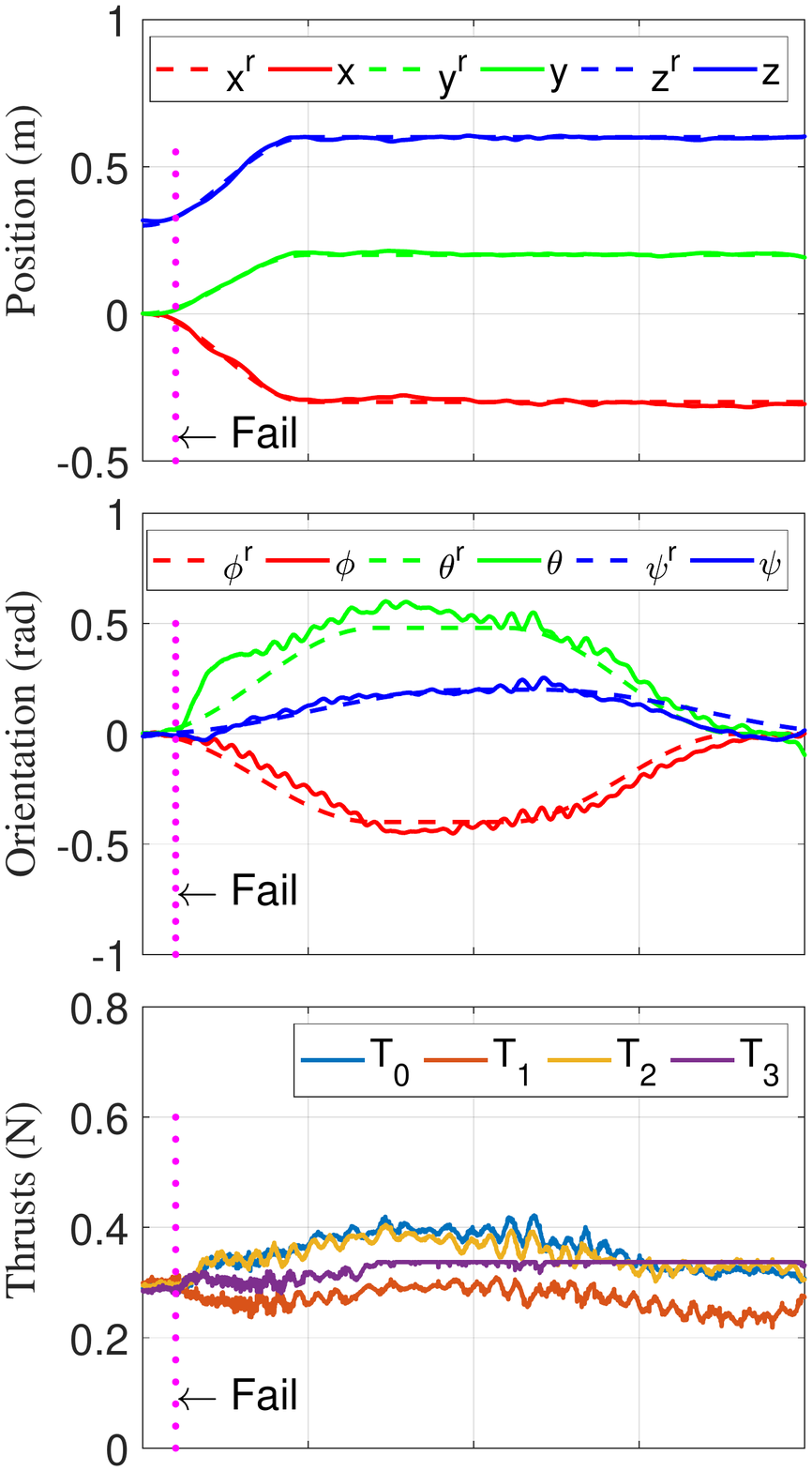}\\
        \includegraphics[width=\linewidth,trim=3.8cm 5.4cm 3.5cm 1.5cm, clip]{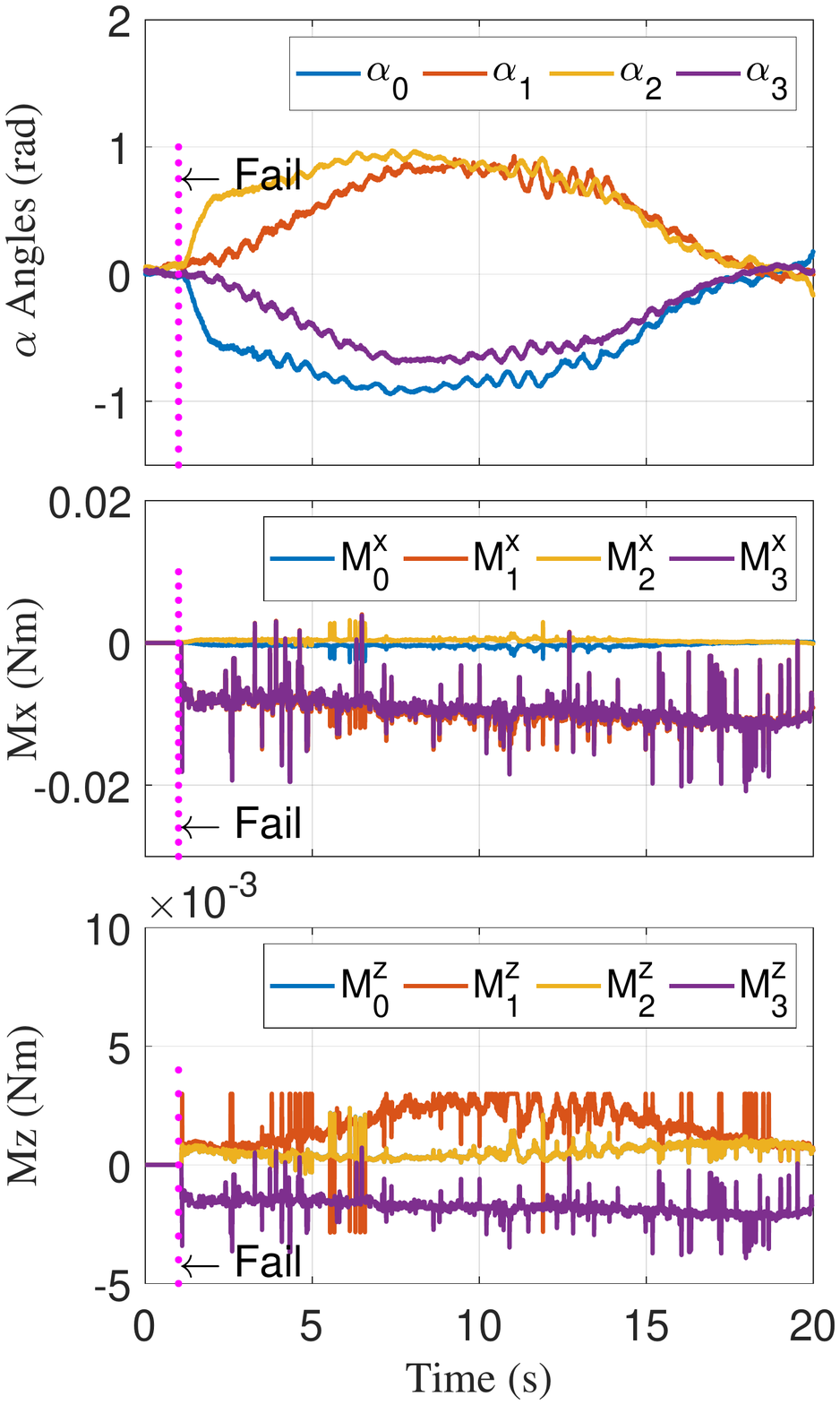}
        \caption{One Fail (S): FTC}
        \label{fig:hinge_opt_1fail_2}
    \end{subfigure}
    \caption{\textbf{Case 2: Trajectory tracking performance when one propeller is failed.} (S stands for the saturated trajectory that $T_3>2\,t_\text{max}$ at some time steps. Reference attitude trajectory with $\phi_{\text{max}}=0.5~rad$, $\theta_{\text{max}}=0.2~rad$, $\psi_{\text{max}}=0.4~rad$.)}
    \label{fig:hinge_1fail_hard}
\end{figure}

As shown in \cref{fig:hinge_1fail}, both controllers maintained the stability of the UAV platform. 
However, for the N+L controller (\cref{fig:hinge_normal_1fail}), disturbance torques $M_3^x$ and $M_3^z$ introduced by low-level adjustment could only be compensated by the integral action in the trajectory tracking controller, leading to the oscillation of the platform. 
In contrast, the \ac{ftc} controller (\cref{fig:hinge_opt_1fail}) incorporated a compensation loop to attenuate disturbance, resulting in improved tracking performance and reduced oscillation for the entire platform.

\paragraph{\textbf{Saturated Trajectory}}
In this test, a more challenging reference trajectory that requires $T_3>2\,t_\text{max}$ at some timesteps (see \cref{fig:hinge_normal_1fail_2}) was designed, activating the low-level saturation constraint of $\mathcal{Q}_3$. Likewise, $\mathcal{P}_0$ of $\mathcal{Q}_3$ was set to fail at $1s$. The tracking performance of the two controllers is plotted in \cref{fig:hinge_1fail_hard}. 

\begin{figure}[t!]    
    \centering
    \begin{subfigure}[b]{0.49\linewidth}
    \centering
        \includegraphics[width=\linewidth,trim=3.8cm 4.2cm 3.5cm 1.7cm, clip]{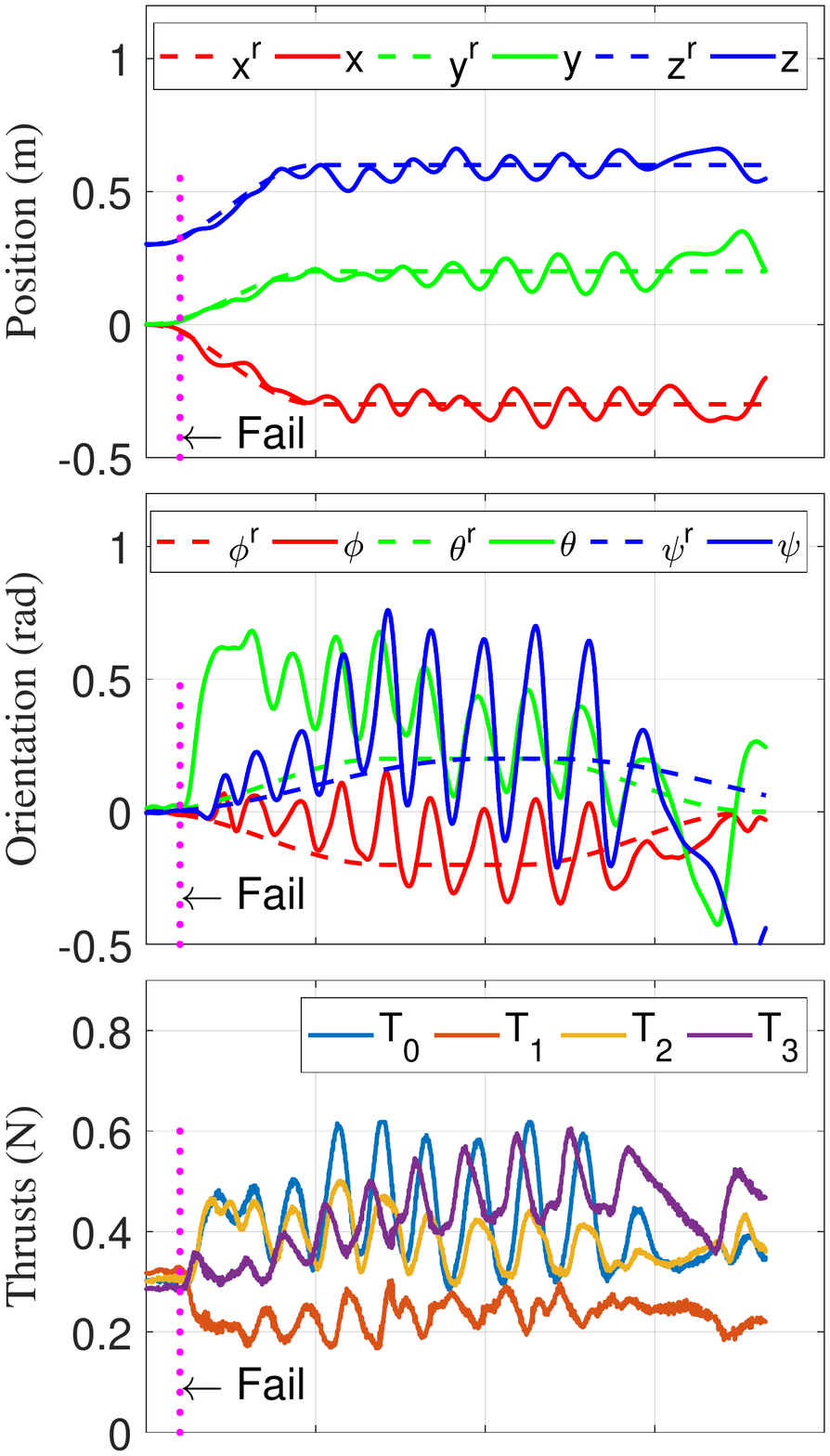}\\
        \includegraphics[width=\linewidth,trim=3.8cm 6cm 3.5cm 1.7cm, clip]{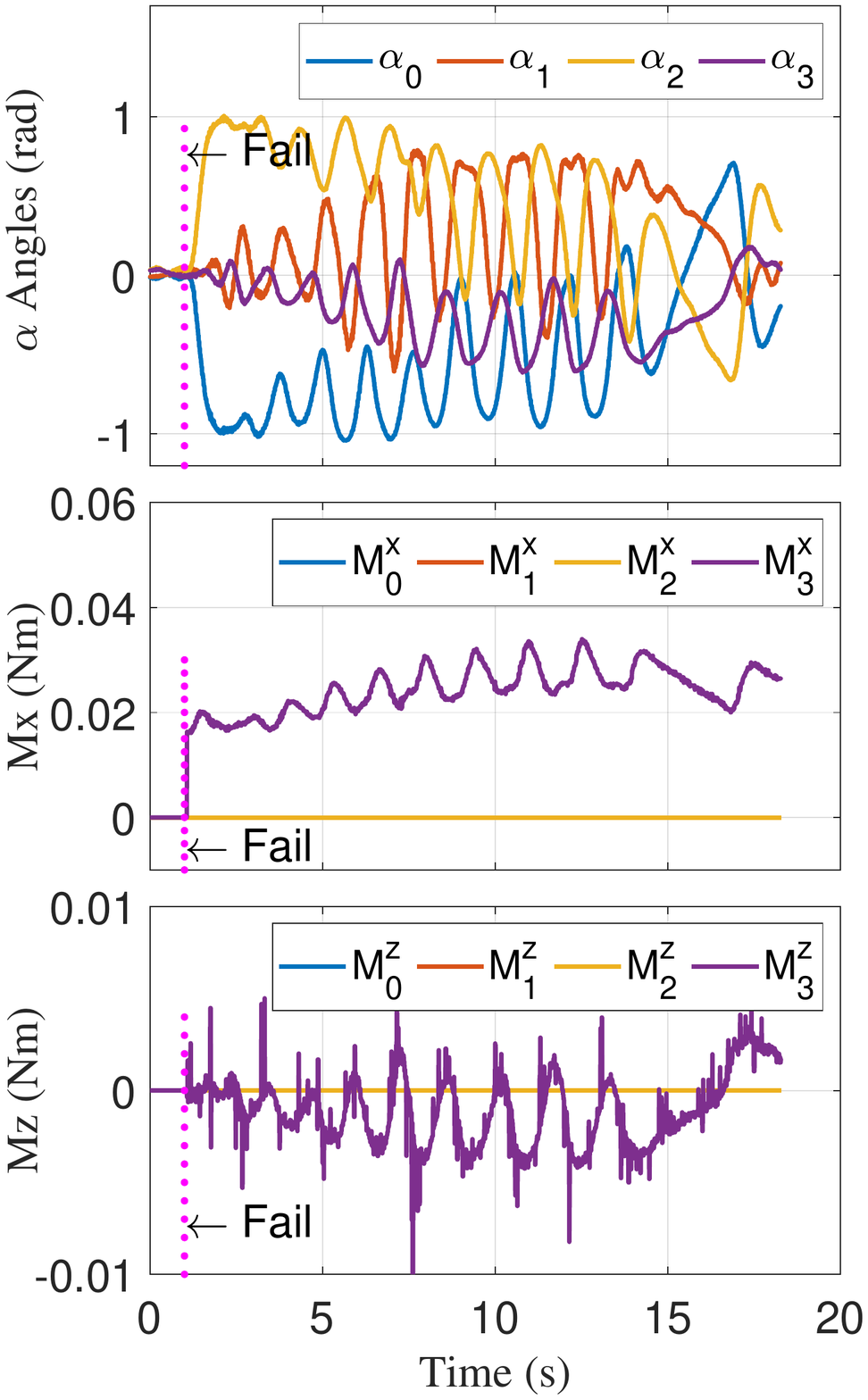}
        \caption{Two Fails: N+L}
        \label{fig:hinge_normal_two_fail}
    \end{subfigure}
    \begin{subfigure}[b]{0.49\linewidth}
    \centering
        \includegraphics[width=\linewidth,trim=3.8cm 4.2cm 3.5cm 1.7cm, clip]{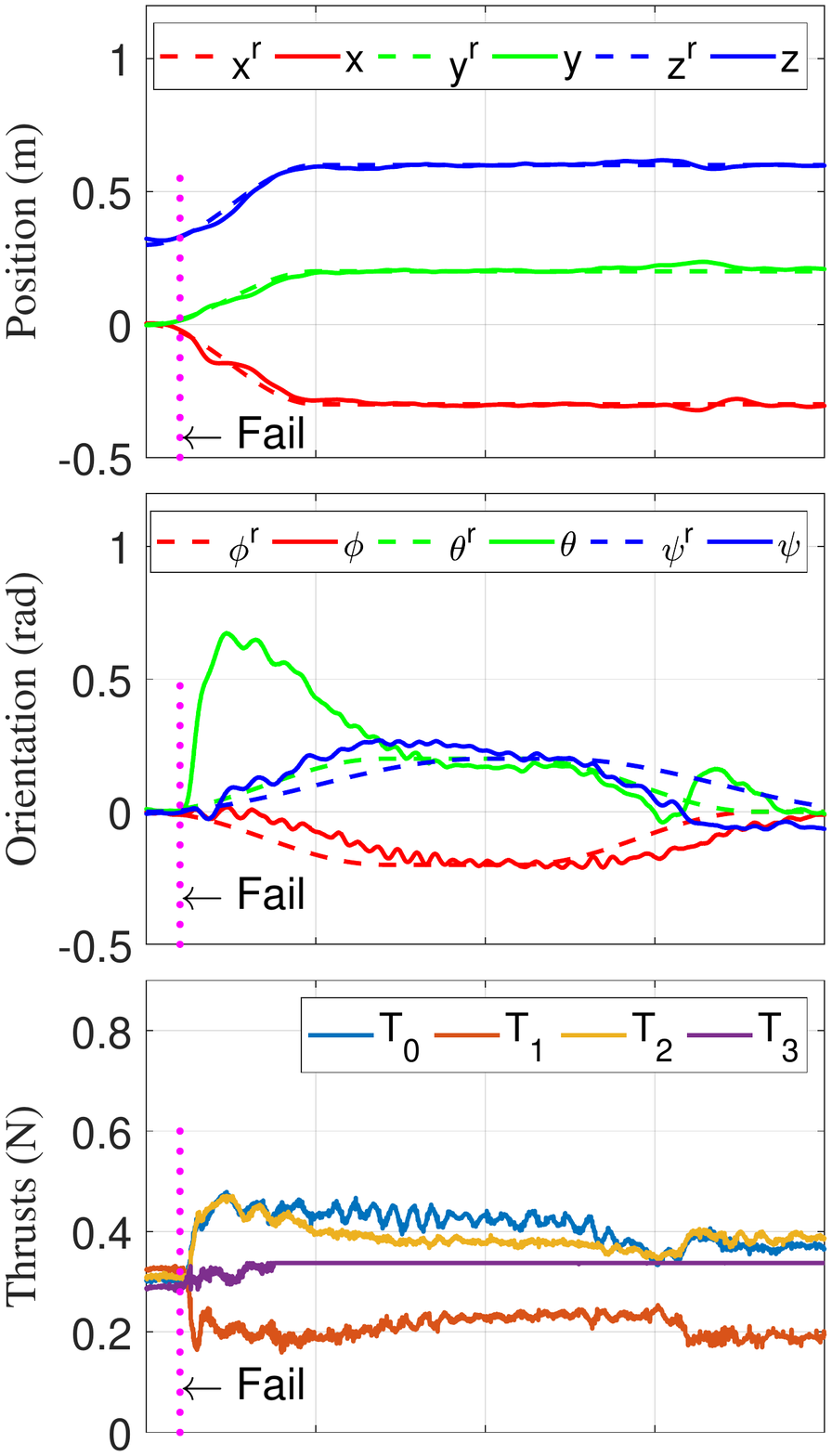}\\
        \includegraphics[width=\linewidth,trim=3.8cm 6cm 3.5cm 1.7cm, clip]{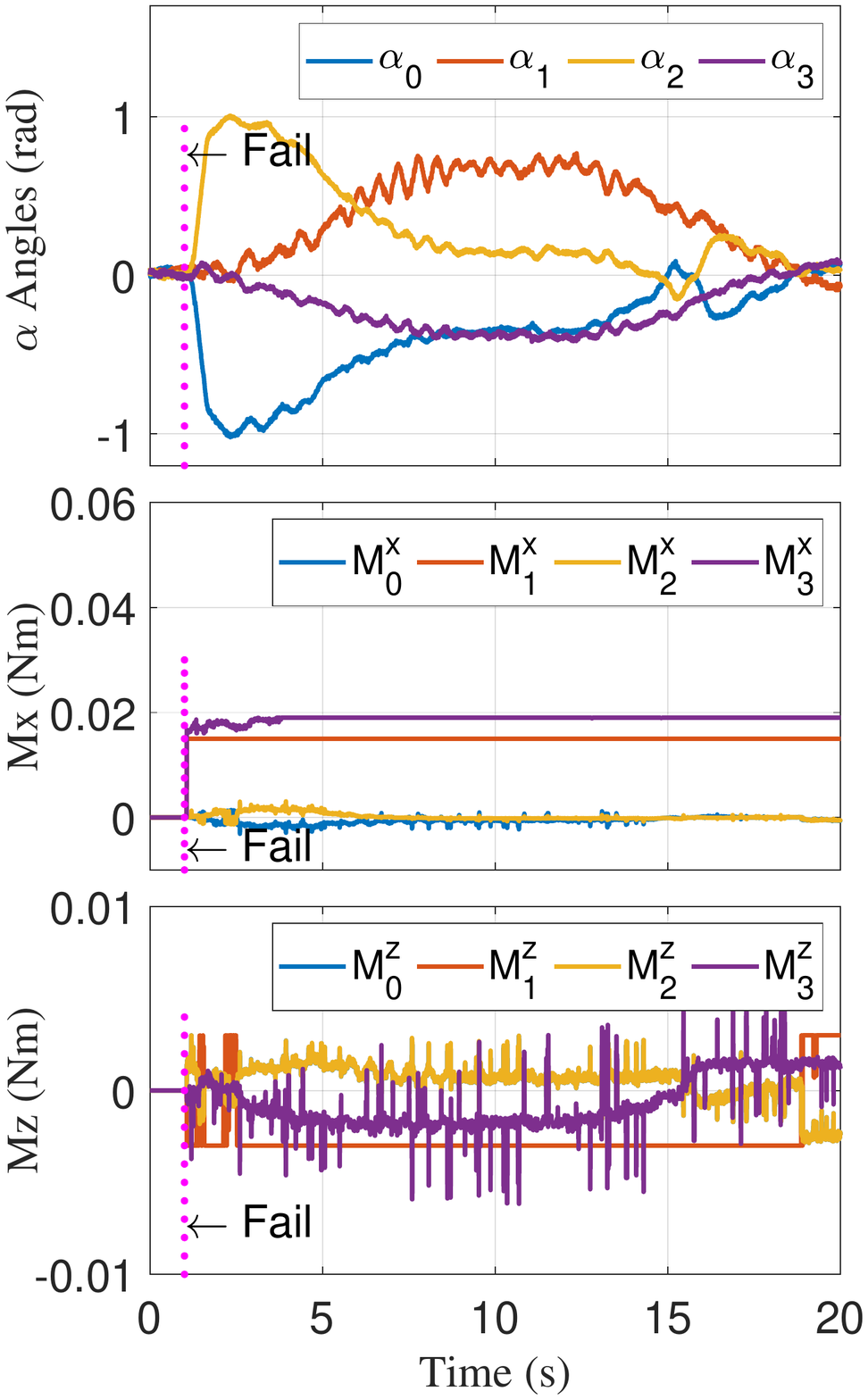}
        \caption{Two Fails: FTC}
        \label{fig:hinge_opt_two_fail}
    \end{subfigure}
    \caption{\textbf{Case 3: Trajectory tracking performance when two propellers failed.} (Reference attitude trajectory with $\phi_{\text{max}}=0.2~rad$, $\theta_{\text{max}}=0.2~rad$, $\psi_{\text{max}}=0.2~rad$.)}
    \label{fig:hinge_two_fail}
\end{figure}

As shown in \cref{fig:hinge_normal_1fail_2}, when the low-level saturation constraint was triggered. the N+L controller caused the UAV platform to become unstable at approximately $7s$. 
Two explanations are given here. First, the nominal allocation strategy did not consider the thrust-saturation constraint, leading to output desired thrust $T_3$ beyond the capability of $\mathcal{Q}_3$. 
Second, the disturbance torques $M_x^3$ and $M_z^3$ generated by the low-level control were transferred to the central frame without compensation. 

For the \ac{ftc} controller (\cref{fig:hinge_opt_1fail_2}), 
the high-level control facilitated thrust distribution
adjustment through the nullspace-based allocation strategy, enabling different saturation values for different quadcopters. 
In addition, the disturbance torques $M_3^x$ and $M_3^z$ can be compensated with the auxiliary inputs. Consequently, position and attitude control remained stable along the entire trajectory.

% ------------------------
\subsubsection{Two Failed Propellers}
In this experiment, $\mathcal{P}_0$ and $\mathcal{P}_1$ of $\mathcal{Q}_3$ began to fail at $1s$ while tracking the reference trajectory. 
The tracking performance of the N+L controller and \ac{ftc} controller are both plotted in \cref{fig:hinge_two_fail}. 
In this scenario, the disturbance torques $M_3^x$ and $M_3^z$ introduced by the low-level controller were larger than in Case 2, making platform stabilization more challenging despite the reference trajectory possessing a relatively smaller tilting attitude compared to Case 2.

As shown in \cref{fig:hinge_normal_two_fail}, 
the entire platform became unstable with the N+L controller. For the \ac{ftc} controller, the platform successfully tracked the reference position and attitude reference trajectories (\cref{fig:hinge_opt_two_fail}). Due to the larger interacting disturbance torques produced by low-level adjustment in this case, the overall performance for both controllers was worse than that observed in Case 2. Some video clips of this experiment are shown in \cref{fig:clips}.

\section{Discussion}
\label{sec:discussion}
This paper proposed an \ac{ftc} to handle certain propeller failure situations for over-actuated \ac{uav} platforms built on quadcopters and passive hinges, which is an increasingly popular category of over-actuated \ac{uav} configurations in the recent years~\cite{nguyen2018novel} for its simplicity in design and prototyping, in addition to the natural attenuation to internal disturbance from propeller drag, momentum and reaction torque, as introduced in~\cite{ruan2023control, ruan2020independent, pi2021simple, yu2021over, su2021compensation}. 

One notable feature of the presented configuration is the existence of high-bandwidth auxiliary inputs $M^x_i$ and $M^z_i$ in the low-level control of $\mathcal{Q}_i$ (\cref{eq: quad input decouple}), as analyzed in~\cite{ruan2023control}. These inputs can either be used to improve the tracking performance as in~\cite{su2021fast}, compensate for the air dynamics as in~\cite{su2022down}, or improve system robustness under certain failure, as demonstrated in this paper. Particularly, in the event of propeller failure, the thrust generator in this type of configuration partially reserves its actuation functionality due to the compensation from auxiliary inputs, so that crashes can be prevented. However, we also witnessed that when the low-level pose regulation controller on the passive joint fails, as shown in \cref{tab:combination}, the uncontrolled relative motion between the quadcopter and mainframe will introduce a huge disturbance to the platform, which shall be further investigated by future research.

The characteristics of the presented configuration also make it befitting as the foundation module of reconfigurable aerial platforms, so that the applications can be extended in multiple directions including flight formation~\cite{su2023flight} or aerial manipulation~\cite{su2023sequential}. The popularity of constructing flying structures with quadcopters reiterates the significance of our platform and the associated control challenges.

\section{Conclusion} \label{sec:conclusion}
In this paper, we addressed the \acf{ftc} method for over-actuated \ac{uav} platforms. Our approach adopts the widely-used hierarchical control architecture and the nullspace-based constrained control allocation we have recently developed for omnidirectional flight.

The logic of FTC within the hierarchical control is presented in the following manner: The low-level control maintains the quadcopter's control of the orientation and thrust whenever possible. The high-level control sets up the maximum thrust available to each quadcopter unit and solves desired commands using the nullspace-based constrained allocation framework.  Uncontrolled disturbances generated by the Bad QC are compensated by the Good QCs whenever within the saturation limit. Both simulation and experimental results, in the cases of one or two propeller failures, have demonstrated FTC's stability and improved trajectory-tracking performance compared to the nominal control method. 

Our FTC analysis and methods encompass all possible combinations of propeller failures and can be applied to other platform configurations with similar low-level propeller actuators and maximum thrust limit setups in the high-level control.  For aerial platforms with fixed rotor angles, the high-level control method can be readily applied with the maximum thrust of failed propellers set to zero.  Finally, in situations where the platform is inevitably failing, graceful crashing is desirable and warrants future investigation.

\section*{Acknowledgments}
The authors would like to thank Dr. Hangxin Liu, Dr. Zeyu Zhang, Dr. Wenzhong Yan, and Dr. Ankur Mehta at UCLA for the figure polishment and the technical assistance with the motion-capture system.
{
%\small
%\setstretch{0.96}
\bibliographystyle{ieeetr}
\bibliography{reference}
}
\vskip -1\baselineskip plus -1fil
\begin{IEEEbiography}[{\includegraphics[width=1in,height=1.25in,clip,keepaspectratio]{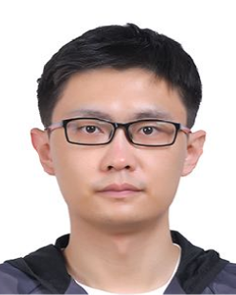}}]{Yao Su}
(Member, IEEE) received the B.S. degree from the School of Mechatronic Engineering, Harbin Institute of Technology in 2016, and the M.S. and Ph.D. degrees from the Department of Mechanical and Aerospace Engineering, University of California, Los Angeles in 2017 and 2021. He is now a research scientist at National Key Laboratory of General Artificial Intelligence, Beijing Institute for General Artificial Intelligence (BIGAI). His research interests include robotics, control, optimization, trajectory planning, and mechatronics.
\end{IEEEbiography}

\vskip -1\baselineskip plus -1fil
\begin{IEEEbiography}[{\includegraphics[width=1in,height=1.25in,clip,keepaspectratio]{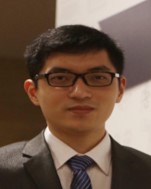}}]{Pengkang Yu}
received the B.Eng. degree in Mechanical Engineering from Hong Kong University of Science and Technology, Hong Kong, in 2016. He received the M.S. degree in 2017 and the Ph.D. degree in 2022 in Mechanical Engineering from the University of California, Los Angeles. His research interests include control, optimization, planning, robotics and mechatronics. 
\end{IEEEbiography}

\vskip -1\baselineskip plus -1fil
\begin{IEEEbiography}[{\includegraphics[width=1in,height=1.25in,clip,keepaspectratio]{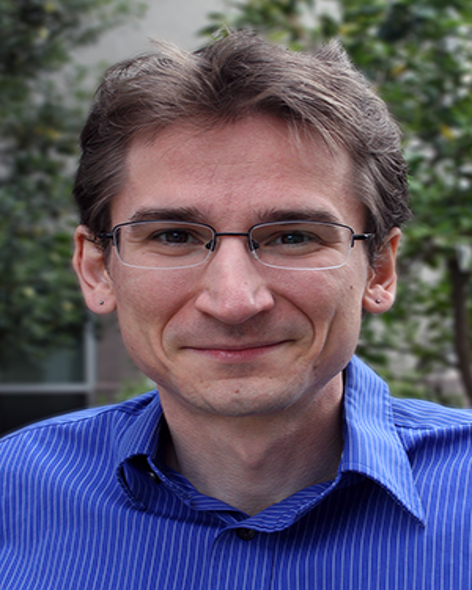}}]{Matthew J. Gerber}
received the B.Eng. degree in Mechanical Engineering from Hong Kong University of Science and Technology, Hong Kong, in 2016. He received the M.S. degree in 2017 and the Ph.D. degree in 2022 in Mechanical Engineering from the University of California, Los Angeles. His research interests include control, optimization, planning, robotics and mechatronics.
\end{IEEEbiography}

\vskip -1\baselineskip plus -1fil
\begin{IEEEbiography}[{\includegraphics[width=1in,height=1.25in,clip,keepaspectratio]{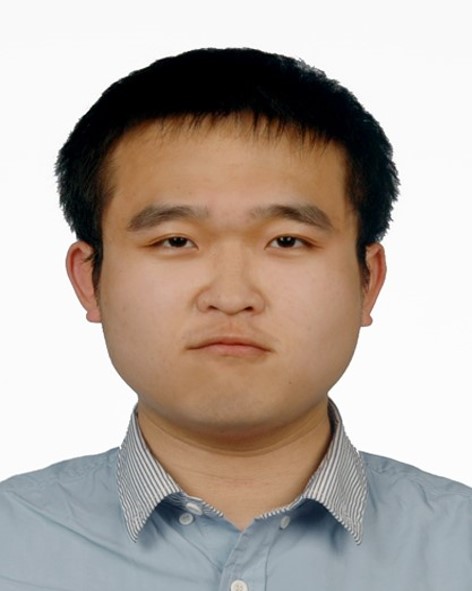}}]{Lecheng Ruan}
received the B.S. honor degree from the School of Mechatronic Engineering, Harbin Institute of Technology in 2015, and the Ph.D. degree from the Department of Mechanical and Aerospace Engineering, University of California, Los Angeles in 2020. He is now affiliated with the National Key Laboratory of General Artificial Intelligence and Peking University. His research interests include control and optimization, mechatronics, robotics, perception and signal processing.
\end{IEEEbiography}

\vskip -1\baselineskip plus -1fil
\begin{IEEEbiography}[{\includegraphics[width=1in,height=1.25in,clip,keepaspectratio]{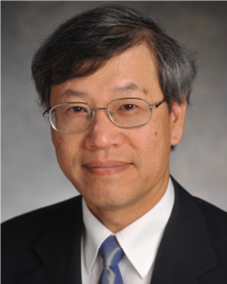}}]{Tsu-Chin Tsao}
(Senior Member, IEEE; Fellow, ASME) received the B.S. degree in engineering from National
Taiwan University, Taipei, Taiwan, in 1981, and the M.S. and Ph.D. degrees in mechanical
engineering from the University of California Berkeley, Berkeley, CA, USA, in 1984 and 1988,
respectively. He is currently a Professor with the Mechanical and Aerospace Engineering Department, University of California, Los Angeles, Los Angeles CA, USA. His research interests include precision motion control, mechatronics, and robotics.
\end{IEEEbiography}
\end{document}